\documentclass[letterpaper]{article} 
\usepackage{aaai2026}  
\usepackage{times}  
\usepackage{helvet}  
\usepackage{courier}  
\usepackage[hyphens]{url}  
\usepackage{graphicx} 
\usepackage{natbib}  
\usepackage{caption} 
\usepackage{algorithm}

\usepackage{amsmath}
\usepackage{amssymb}
\usepackage{algpseudocode}
\usepackage{booktabs}
\usepackage{colortbl}
\usepackage{multirow}
\usepackage{subcaption} 
\usepackage{tabularx}
\usepackage{xspace}
\usepackage{xcolor}
\usepackage{adjustbox}
\usepackage{bm}

\def\onedot{.}
\def\eg{\emph{e.g}\onedot}
\def\ie{\emph{i.e}\onedot}
\def\viz{\emph{viz}\onedot}
\def\vh{{\bm{h}}}

\newcommand{\sname}{LfU\xspace}
\def\vtheta{{\bm{\theta}}}
\def\vx{{\bm{x}}}
\def\vy{{\bm{y}}}
\def\vd{{\bm{d}}}
\def\vz{{\bm{z}}}
\def\sR{{\mathbb{R}}}

\usepackage{newfloat}
\usepackage{listings}
\DeclareCaptionStyle{ruled}{labelfont=normalfont,labelsep=colon,strut=off} 
\floatstyle{ruled}
\newfloat{listing}{tb}{lst}{}
\floatname{listing}{Listing}
\title{Learning from the Undesirable: \\ Robust Adaptation of Language Models without Forgetting}
\author {
    Yunhun Nam\textsuperscript{\rm 1},
    Jaehyung Kim\textsuperscript{\rm 2},
    Jongheon Jeong\textsuperscript{\rm 1}
}
\affiliations {
    \textsuperscript{\rm 1}Korea University,\\
    \textsuperscript{\rm 2}Yonsei University\\
    \{yh0326, jonghj\}@korea.ac.kr, 
    jaehyungk@yonsei.ac.kr

}

\usepackage{bibentry}

\begin{document}

\maketitle

\begin{abstract}
Language models (LMs) are often adapted through supervised fine-tuning (SFT) to specialize their capabilities for downstream tasks. However, in typical scenarios where the fine-tuning data is limited, \eg, compared to pre-training, SFT can lead LMs to overfit, causing them to rely on spurious patterns within the target task or to compromise other broadly useful capabilities as a side effect of narrow specialization.
In this paper, we propose \emph{Learning-from-the-Undesirable} (\sname), a simple yet effective 
regularization scheme for SFT to mitigate overfitting issues when fine-tuning LMs with limited data. Specifically, we aim to regularize the fine-tuning process to favor solutions that are resilient to ``undesirable'' model updates, \eg, gradient ascent steps that steer the model toward undesirable behaviors. To this end, we propose a novel form of consistency regularization that directly aligns internal representations of the model with those after an undesirable update. By leveraging representation-level data augmentation through undesirable updates, \sname effectively promotes generalization under limited data. Our experiments on diverse LM downstream tasks show that \sname serves as an effective prior that enhances adaptability while preserving pretrained knowledge.  For example, our LM from \sname achieves a $16.8\%$ average improvement on math tasks compared to vanilla SFT on the same dataset, where the latter even leads to degraded performance on those tasks. 
Furthermore, \sname exhibits improved robustness to prompt variations, \eg, yielding a $92.1\%$ lower standard deviation in output performances compared to SFT, highlighting its versatile effects.
\end{abstract}

\begin{links}
    \link{Code}{https://github.com/yunpal/LfU}
\end{links}

\section{Introduction}
{Language models (LMs) \cite{touvron2023llama,mann2020language,chatgpt,openai2023gpt4,claude,geminiteam2023gemini} have recently emerged as strong backbones for various downstream tasks, thanks to their unprecedented capabilities in understanding natural language and encoding vast world knowledge:} \eg, in question answering \cite{lewis2020retrieval}, safeguard modeling \cite{inan2023llama}, code generation \cite{roziere2023code}, API call generation \cite{patil2024gorilla} and reward modeling \cite{rafailov2023direct}. {\emph{Supervised fine-tuning} (SFT) is currently a standard approach for adapting LMs to specific tasks; \ie, by fine-tuning them on curated input-output pairs that reflect the desired behaviors. Ultimately, SFT aims to reliably steer model behavior towards specific tasks while effectively preserving the general knowledge and reasoning capabilities of LMs.}

\begin{figure}[t]  
\centering
\includegraphics[width=0.5\textwidth]{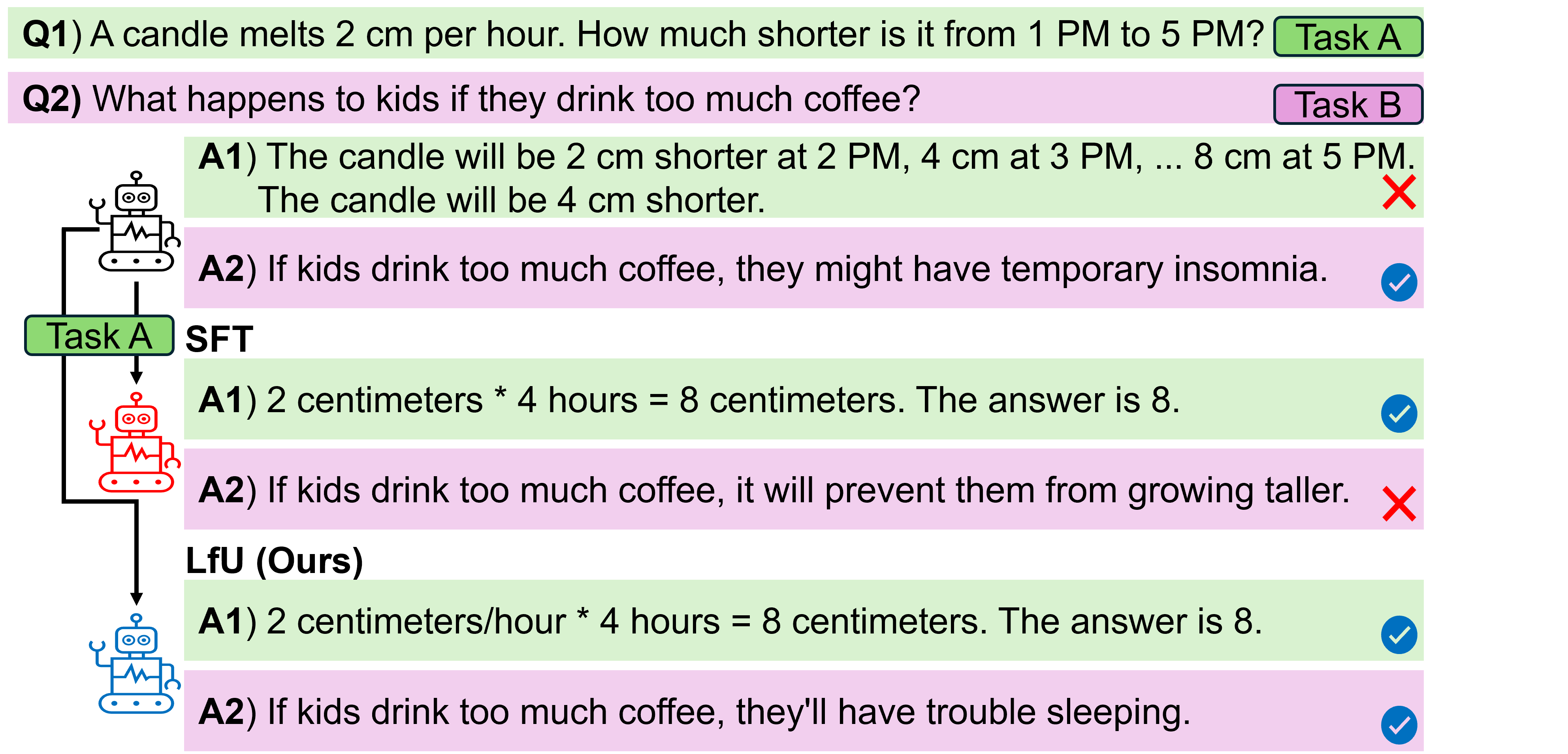}
\caption{Illustration of forgetting in SFT vs. preservation in \sname (Ours): Fine-tuning via SFT on Task A causes the model to forget prior knowledge related to Task B. In contrast, \sname successfully learns Task A while preserving the prior knowledge about Task B.}
\label{fig:concept}
\end{figure}

\begin{figure*}[t]
\centering
\begin{subfigure}[b]{0.31\textwidth}
    \centering
    \includegraphics[width=\textwidth]{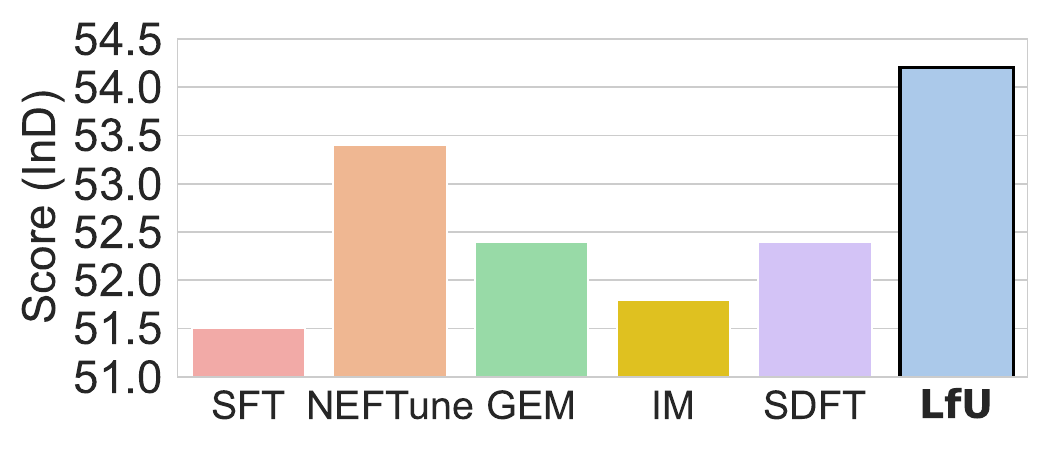}
    \caption{GSM8k $\rightarrow$ Math (In-domain)}
    \label{fig:single_base}
\end{subfigure}
\hfill
\begin{subfigure}[b]{0.31\textwidth}
    \centering
    \includegraphics[width=\textwidth]{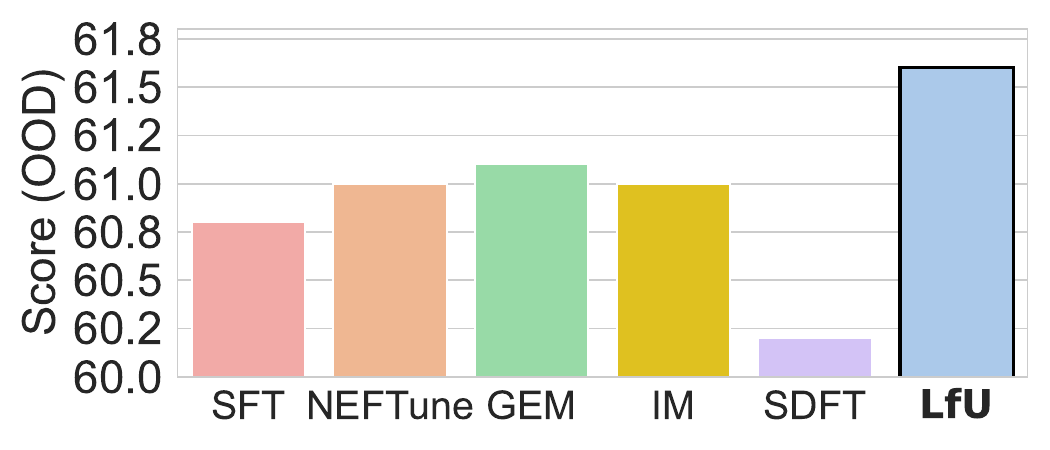}
    \caption{GSM8k $\rightarrow$ All (Out-of-domain)}
    \label{fig:single_ins}
\end{subfigure}
\hfill
\begin{subfigure}[b]{0.31\textwidth}
    \centering
    \includegraphics[width=\textwidth]{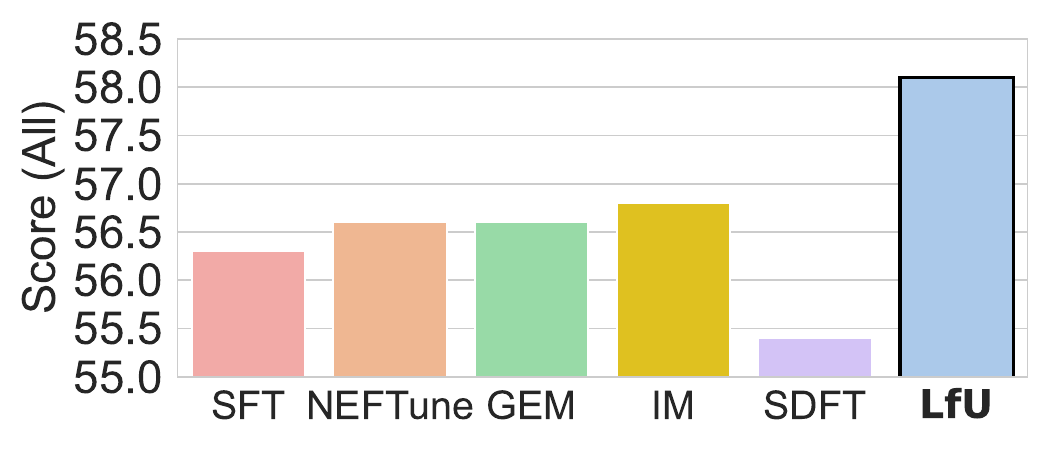}
    \caption{Alpagasus Dolly 3k $\rightarrow$ All}
    \label{fig:multi}
\end{subfigure}
\caption{Performance comparison between baselines and \sname fine-tuned on Llama-3.1-8B. (a) and (b) show results fine-tuned on GSM8k and evaluated on in-domain and out-of-domain data, respectively. These results indicate that while prior methods struggle to adapt to in-domain examples and lead to only marginal improvements on out-of-domain data, \sname consistently achieves the best performance in both cases. (c) presents results from fine-tuning on the multitask dataset Alpagasus Dolly 3k, with evaluation across all tasks. \sname achieves the best overall performance.}
\label{fig:perf_graph}
\end{figure*}

In practice, SFT often suffers from overfitting, which not only forgets prior knowledge but also increases vulnerability to adversarial fine-tuning. As shown in figure~\ref{fig:concept}, the model adapted to a downstream task often forgets prior knowledge. Furthermore, they exhibit high sensitivity to prompt variations during inference \cite{reynolds2021prompt, zhu2023promptbench}, making their behavior unstable across semantically equivalent inputs. In addition, SFT suffers from fragile alignment, as it tends to overfit to superficial refusal patterns rather than learning robust safety behaviors. As demonstrated by \cite{qi2024safety}, only a few steps of adversarial fine-tuning are sufficient to undo these behaviors and elicit harmful outputs.

To mitigate these limitations, recent efforts have focused on mitigating overfitting during SFT \cite{jain2023neftune,shi2024instruction,li2025preserving,yang2024self}. 
For example, \citet{jain2023neftune} proposes injecting small noise into the input token embedding vectors during the forward pass of training. 
Other approaches focus on refining the standard cross-entropy objective in SFT. \citet{shi2024instruction} proposes Instruction Modelling (IM), which applies the loss function not only over the output but also over the instruction. 
This encourages the model to pay closer attention to the instruction and improves generalization by mitigating overfitting. \citet{li2025preserving} proposes Game-theoretic Entropy Maximization (GEM), which preserves output diversity by modeling supervised learning as a distribution matching game with entropy regularization. As shown in Figure~\ref{fig:perf_graph}, these methods lead to only marginal improvements on both in-domain (InD) and out-of-domain(OOD) examples. In a complementary direction, \citet{yang2024self} proposes rewriting the outputs of the training data to resemble those of the pre-trained model, thereby better aligning the training data distribution with the original output distribution. However, Figure~\ref{fig:perf_graph} shows that it requires an instruction-tuned model to begin with, making it incompatible with untuned vanilla models.

\paragraph{Contributions}

In this paper, we propose a simple yet effective method to mitigate overfitting, coined \emph{Learning-from-the-Undesirable} (\sname). Overfitting often arises in SFT due to limited training data, where the model tends to memorize spurious patterns. A common strategy to alleviate this issue is data augmentation \cite{wei2019eda,feder2023data}, which improves generalization by exposing the model to diverse variations of inputs. Inspired by this, we leverage the effect of data augmentation at the representation level by introducing undesirable updates to generate diverse internal representations for the same input. The key idea of \sname is to simulate undesirable updates by constructing an auxiliary model and applying a single gradient ascent step that shifts the model in an undesirable direction. To do this, we augment the original model with trainable components, \eg, LoRA parameters or representation steering vectors, and compute the gradient of the SFT objective with respect to these components. Using this gradient, we can get an undesirable model. Then, \sname regularizes the training process by enforcing consistency between the internal representations of the original and undesirable models, thereby encouraging the model to maintain stable internal representations. As shown in Figure~\ref{fig:perf_graph}, our undesirable update strategy provides more effective augmentation and leads to better generalization compared to simple perturbation methods, such as adding Gaussian noise to input embeddings as done in NEFTune~\cite{jain2023neftune}. 

Extensive experiments across a wide range of downstream tasks and language models demonstrate the effectiveness of \sname. For example, when fine-tuning Llama-3.1-8B on GSM8k, \sname improves in-domain performance by up to $16.8\%$ over SFT, outperforming all existing baselines in both in-domain and out-of-domain datasets. Moreover, \sname exhibits strong resilience to prompt variations, reducing standard deviation in output performances by $92.1\%$ compared to SFT and improving average accuracy. Furthermore, in a few steps of adversarial fine-tuning scenarios, \sname significantly reduces the Attack Success Rate (ASR), achieving up to a $45.0\%$ lower ASR on harmful datasets over SFT.

\section{Related works}
\label{sec:related}

\paragraph{Overfitting issues in SFT} 
It has been observed that SFT {tends to overfit limited data, causing} forgetting and hallucinations \cite{kotha2023understanding, gekhman2024does, lin2024flame}; these problems can cause undesirable shifts during downstream adaptation \cite{luo2023empirical, jiang2025unlocking, qi2023fine}. To address this challenge, several studies have explored data augmentation \cite{wei2019eda,feder2023data}, \eg, including the use of synthetic data~\cite{ba2024fill}. However, due to limited semantic diversity, synthetic data alone often leads to hallucinated or biased outputs and may not offer a fundamental solution \cite{guo2023curious,shumailov2024ai, rogulsky2024effects, fang2024bias}. Another direction develops more principled training methods to better support generalization. For instance, \citet{jain2023neftune,yadav2023symnoise} tackle the challenge by injecting noise into embeddings during training. {\citet{shi2024instruction} proposed to apply the cross-entropy loss to both outputs and instructions, and \citet{li2025preserving} aim to encourage the output diversity.} Another recent study aims to reduce distributional shift by fine-tuning the model on their own generated responses \cite{yang2024self}. In this paper, we aim to enhance generalization of SFT through a new consistency regularization and update dynamics.

\paragraph{Consistency regularization}

As an alternative to data augmentation, \emph{consistency regularization} \cite{sajjadi2016regularization} offers a principled approach to improve generalization by enforcing stable predictions under small perturbations. It has been widely studied across various areas, including semi-supervised learning \cite{xie2020unsupervised, sohn2020fixmatch, tang2023consistency}, robust learning \cite{zhang2019theoretically, jeong2020consistency, tack2022consistency}, generative modeling \cite{zhang2020Consistency, ni2025noise}, and bias mitigation \cite{wang2025debiased}. In this paper, \sname enforces internal representation consistency to mitigate overfitting by stabilizing the model against corruptions of itself, which we refer to as \emph{undesirable updates}.

\paragraph{Learning from update dynamics}
Understanding and leveraging the dynamics of parameter updates has served as a powerful strategy for improving model generalization. 
One prominent example is meta-learning \cite{thrun1998learning,rajeswaran2019meta}, which trains models to quickly adapt to unseen tasks rather than optimizing for a single fixed objective. In particular, meta-learning simulates gradient-based updates on diverse tasks during training \cite{finn2017model,li2017meta}, enabling the model to learn a parameter initialization that can be efficiently fine-tuned for new tasks. Sharpness-aware training, such as SAM \cite{foret2020sharpness,oikonomou2025sharpness,tahmasebi2024universal}, takes a different approach by performing small gradient ascent steps to locate flatter minima, thereby reducing overfitting and improving robustness. While prior methods leverage update dynamics mainly for in-domain improvements, LfU operates at the representation level to achieve stable and generalizable training.

\section{Method}

We denote the language model by $p_{\vtheta}$, where $\vtheta$ represents the model parameters. 
Given an input $\vx$, the model generates an output $\vy$ by sampling each token $\vy_t \sim p_\vtheta(\cdot \mid \vx, \vy_{<t})$, where $t$ denotes the index of the output token. 
SFT adapts a language model to downstream tasks by minimizing the negative log-likelihood of the output tokens. 
To be specific, given a dataset $\mathcal{D}$ of input-output pairs $(\vx, \vy)$, where an output of length $T$, the SFT objective is defined as:
\begin{equation}
\ell_{\text{SFT}} (\vtheta) =
\mathop{\mathbb{E}}\limits_{(\vx, \vy) \sim \mathcal{D}} \left[
- \frac{1}{T}\sum_{t=1}^{T} \log p_\vtheta(\vy_t \mid \vx, \vy_{<t})
\right].
\label{eq:sft}
\end{equation}
SFT has been applied to adapt LMs on specific datasets, but it often induces narrow response patterns and degrade generalization to other unrelated tasks. Although subsequent alignment stages, \eg, via RL \cite{ouyang2022training,kirk2023understanding}, can help mitigating the issue, it opens up new challenges in specifying rewards and requires significant compute. In this paper, we aim to improve SFT in a way that makes it generalizable, so that the model can preserve its original capability even after specializing to a certain task.

\begin{figure}[t]
\centering
\begin{subfigure}[b]{0.48\linewidth}
    \centering
    \includegraphics[width=\linewidth]{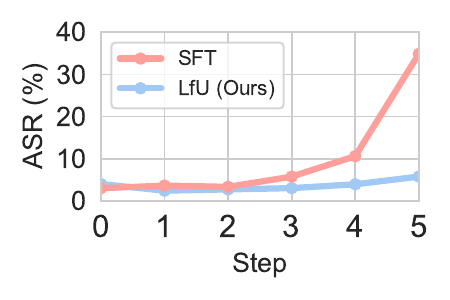}
    \caption{HEx-PHI}
    \label{fig:hex_adv}
\end{subfigure}
\hfill
\begin{subfigure}[b]{0.48\linewidth}
    \centering
    \includegraphics[width=\linewidth]{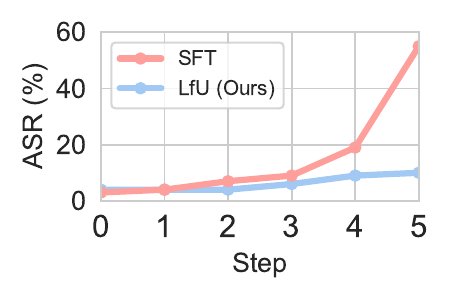}
    \caption{PureBad}
    \label{fig:purebad}
\end{subfigure}
\caption{
Attack Success Rate (ASR) on (a) HEx-PHI and (b) PureBad after a few steps of adversarial fine-tuning on BeaverTails \cite{ji2023beavertails}. We first align a Llama-3.1-8B via SFT (or \sname) with the harmless subset of BeaverTails, and then continue fine-tuning the model on the harmful subset (of BeaverTails) using SFT.}
\label{fig:sft_bad}
\end{figure}

\begin{figure*}[t]  
\centering
\includegraphics[width=1.0\textwidth]{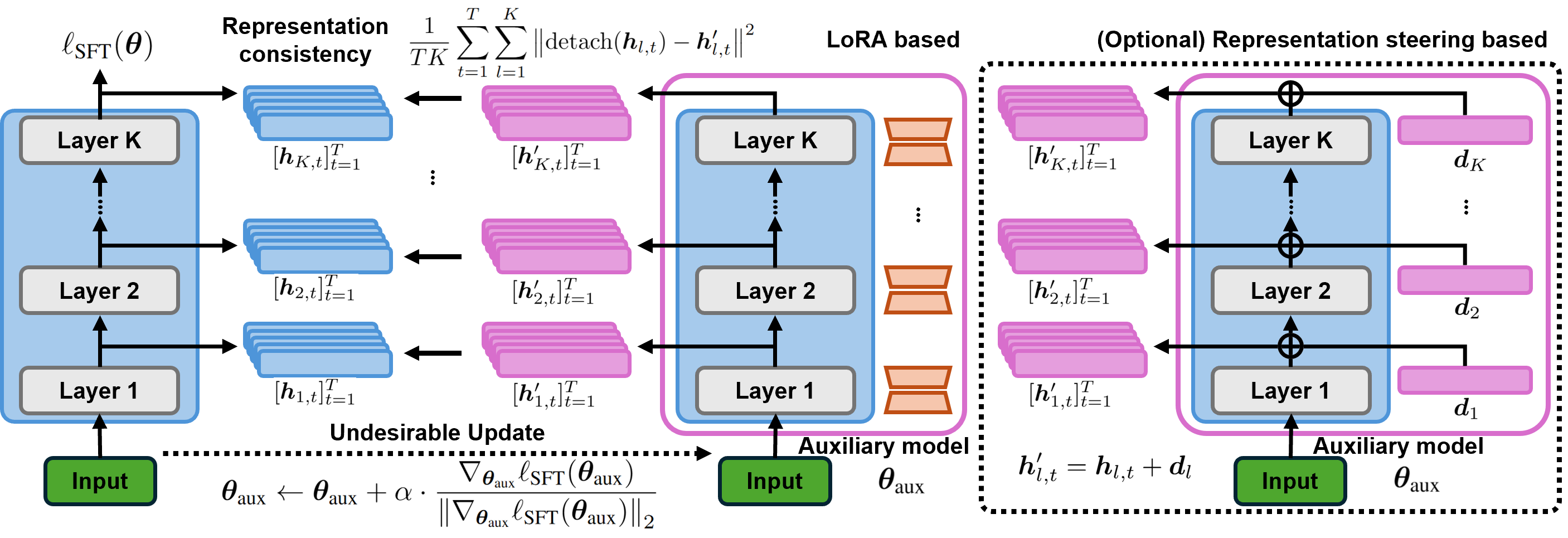}
\caption{
Overview of LfU: LfU promotes stable internal representations by enforcing consistency in internal representations between the original model $\vtheta$ and auxiliary model $\vtheta_{\text{aux}}$ that is optimized one step to induce undesirable behaviors. The auxiliary model $\vtheta_{\text{aux}}$ is constructed by adding additional components to the original parameters $\vtheta$, using either (1) a LoRA based method, where trainable low-rank matrices are added to each layer or (2) a representation steering based method, where a learnable steering vector is added to the internal representation at each layer. A gradient ascent step is then performed on the additional components by computing the gradient of the SFT objective with respect to $\vtheta_{\text{aux}}$. The consistency loss is defined as the Mean Squared Error (MSE) between the internal representations of the original and auxiliary models across all layers.}
\end{figure*}

\paragraph{Motivation}
\label{sec:adver}
We are motivated by the observations in \citet{qi2024safety} that SFT tends to overfit when applied for safety, \ie, to output short refusal response to harmful instructions. As illustrated in Figure~\ref{fig:sft_bad}, 
we observe that the behavior of such an SFT-tuned model, say $p_{\text{SFT}}$, can be easily altered by just a few steps of ``adversarial'' SFT on harmful datasets. To address this, we propose a learning strategy that enhances generalization by simulating undesirable behaviors and encouraging the model to remain robust against them. As shown in Figure~\ref{fig:sft_bad}, our approach remains robust even after a few steps of adversarial fine-tuning.\footnote{The detailed experimental setup is provided in Appendix~\ref{ap:experimental details}.}

\subsection{Approach}

We propose \emph{Learning-from-the-Undesirable} (\sname), a regularization method that encourages the model to generalize better by reducing overfitting. The key idea is to simulate undesirable behaviors within the model and guide it to maintain stable internal representations under such conditions.

\paragraph{One step towards the undesirable}
To encourage stable representations, we simulate a single optimization step that shifts the model parameters toward undesirable directions. To this end, we define an auxiliary model $\vtheta_{\text{aux}}$, which augments the parameters $\vtheta$ with additional components. Specifically, the loss is computed with Eq.~\ref{eq:sft}, and we compute its gradient with respect to the $\vtheta_{\text{aux}}$ to determine the direction of perturbation. A single step of gradient ascent is then applied to update the auxiliary components, as follows:

\begin{equation}
\vtheta_{\text{aux}} \leftarrow \vtheta_{\text{aux}} + \alpha \cdot \frac{
\nabla_{\vtheta_{\text{aux}}} \ell_{\text{SFT}}(\vtheta_{{\text{aux}}})
}{\left\| \nabla_{\vtheta_{\text{aux}}} \ell_{\text{SFT}}(\vtheta_{{\text{aux}}}) \right\|_2},
\label{eq:perturbed_para}
\end{equation}
where $\alpha $ is a step size controlling the magnitude of the perturbation.
This perturbation intentionally induces undesirable behavior.

We next extract internal representations from both the original model $\vtheta$ and the auxiliary $\vtheta_{\text{aux}}$. Given the $l_{th}$ layer of an LM, we define the function  {$M^{(l)}(\cdot; \vtheta)$} to return the internal representations at layer $l$ under parameters $\vtheta$. Similarly, we obtain the function {$M^{(l)}(\cdot; \vtheta, \vtheta_{\text{aux}})$} to return the undesirable representations.
We formally define the representations of $l_{th}$ layer  as follows:
\begin{equation}
\vh_{l,t} = M^{(l)}(\vx, \vy_{<t}; \vtheta),
\end{equation}
\begin{equation}
\vh'_{l,t} = M^{(l)}(\vx, \vy_{<t}; \vtheta, \vtheta_{\text{aux}}),
\end{equation}
where $\vh_{l,t}, \vh'_{l,t} \in \sR^{d}$ denote the representations from the original and auxiliary models, respectively and $d$ is the dimensionality of the representations.

\paragraph{Additional components to make auxiliary model}
There are two ways to construct $\vtheta_{\text{aux}}$: the LoRA method and the representation steering method. 
The LoRA based method leverages Low Rank Adaptation (LoRA) \cite{hu2022lora}. We define the model parameters $\vtheta$ with $K$ layers as $\vtheta = [\vtheta_1, \vtheta_2, \dots, \vtheta_K]$, where $\vtheta_l$ denotes the parameters of the $l$-th layer. Each layer $\vtheta_l$ is augmented with a trainable low-rank matrix $\mathrm{LoRA}_l$, forming $\vtheta_{\text{aux}} = [\vtheta_1 + \mathrm{LoRA}_1, \dots, \vtheta_K + \mathrm{LoRA}_K]$. Only the $\mathrm{LoRA}$ terms are updated, while $\vtheta$ remains frozen. 
 The representation steering based method perturbs internal representations instead of parameters. At each layer $l$, a learnable steering vector $\vd_l$ is added to the internal representation: $\vh'_{l,t} = \vh_{l,t} + \vd_l$. This shifts internal representations in an undesirable direction without modifying $\vtheta$, making it a computationally efficient alternative.

\paragraph{One step towards the desirable}
To encourage stability against undesirable perturbation, we enforce consistency between the model’s internal representations before and after applying the perturbation, which was previously defined. We define the consistency regularization as:
\begin{multline}
    \ell_{\text{cons.}}(\vtheta,\vtheta_{\text{aux}}) \\ 
    =
\mathop{\mathbb{E}}\limits_{(\vx, \vy) \sim \mathcal{D}} \Bigg[ \,
\frac{1}{TK} \sum_{t=1}^{T} \sum_{l=1}^{K}
\left\| \mathrm{detach}(\vh_{l,t}) - \vh'_{l,t} \right\|^2
\Bigg] ,
\end{multline}
where $ \mathrm{detach}(\cdot)$ indicates that gradients do not flow. This design helps stabilize the internal representation $\vh_{l,t}$ by encouraging $\vh'_{l,t}$ to remain close to the fixed target. We incorporate the consistency objective into the overall training loss by combining it with the SFT objective. The resulting objective of LfU is defined as:
\begin{equation}
\ell_{\text{LfU}}(\vtheta,\vtheta_{\text{aux}}) =\ell_{\text{SFT}}(\vtheta) + \lambda \cdot \ell_{\text{cons.}}(\vtheta,\vtheta_{\text{aux}}),
\end{equation}
where $\lambda$ is a hyperparameter.
Detailed pseudocodes to implement LfU are provided in Appendix~\ref{ap:pseudo_lfu}.

\begin{table}[t]
\centering
\resizebox{1.0\linewidth}{!}{%
\begin{tabular}{l|cccc|c}
\multicolumn{1}{c|}{} & \textbf{Math (3)} & \textbf{Knowl. (4)} & \textbf{Reason. (2)} & \textbf{Helpful. (2)}  & \textbf{Rank} \\
\midrule
\rowcolor[rgb]{.816, .816, .816} \multicolumn{6}{l}{\textbf{Llama-3.1-8B}} \\
\midrule
Vanilla & 37.9 & 72.9 & 66.1 & 44.0 & -- \\
\midrule
SFT & 51.5  & 73.3  &65.1 & 44.1  & 4.8 \\
NEFTune  & \underline{53.4}  & 73.3  & \underline{65.3}  & 44.3  & \underline{3.0}  \\
GEM   & 52.4  & 73.0     & \underline{65.3}  & \underline{45.1} & 3.5 \\
IM   & 51.8  & 73.1  & \textbf{65.6} & 44.4  & 3.5 \\
SDFT  & 52.4  & \underline{73.4}  & 65.0     & 42.1  & 4.3 \\
\midrule
\textbf{\sname}& \textbf{54.2} & \textbf{73.5} & \textbf{65.6} & \textbf{45.7}  & \textbf{1.0}  \\
\midrule
\rowcolor[rgb]{.816, .816, .816} \multicolumn{6}{l}{\textbf{Llama-3.1-8B-Instruct}} \\
\midrule
Vanilla & 60.5 & 74.0 & 81.0 & 69.9 & -- \\
\midrule
SFT & 65.7  & 73.6  & 80.5  & 67.6  & 4.0  \\
NEFTune & \underline{66.0}     & 73.6  & 80.7  & 67.3  & 4.3 \\
GEM  & 65.0     & 73.8  & \underline{81.3}  & 67.6  & 3.8 \\
IM  & 63.9  & \textbf{74.1} & \textbf{81.5} & \underline{68.5}  & \underline{2.5} \\
SDFT  & 65.5  & \underline{74.0}     & 79.8  & 67.4  & 4.3 \\
\midrule
\textbf{\sname} & \textbf{66.7} & \underline{74.0}     & \textbf{81.5} & \textbf{69.0 } & \textbf{1.3} \\
\midrule
\rowcolor[rgb]{.816, .816, .816} \multicolumn{6}{l}{\textbf{Llama-2-7B}} \\
\midrule
Vanilla & 17.1  & 66.4  & 60.4  & 41.0     & -- \\
\midrule
SFT  & 35.9  & 65.9  & 60.0     & 40.5  & 3.8 \\
NEFTune  & \underline{36.1}  & 66.2  & 60.1  & 40.4  & \underline{3.3} \\
GEM & 35.9  & \underline{66.4}  & 59.9  & \underline{40.6}  & \underline{3.3} \\
IM & 32.2  & \underline{66.4}  & \textbf{61.0 } & 39.7  & 3.8 \\
SDFT  & 33.6  & 65.6  & 60.0     & 40.1  & 5.0  \\
\midrule
\textbf{\sname} & \textbf{37.0 } & \textbf{66.7} & \underline{60.2}  & \textbf{40.9} & \textbf{1.3} \\
\midrule
\rowcolor[rgb]{.816, .816, .816} \multicolumn{6}{l}{\textbf{Mistral-7B-v0.3}} \\
\midrule
Vanilla & 14.7  & 72.9  & 64.9  & 42.5  & -- \\
\midrule
SFT & 48.2  & 72.8  & 64.8  & \underline{43.1}  & 4.5 \\
NEFTune & \underline{50.3}  & 72.9  & 65.4  & 43.0     & 3.5 \\
GEM  & 49.9  & \underline{73.1}  & \underline{65.6}  & 42.9  & \underline{3.0}  \\
IM  & 43.0     &\underline{73.1}  & 65.4  & 42.3  & 4.8 \\
SDFT  & 46.4  & \textbf{73.5}  & \textbf{65.9} & 42.8  & \underline{3.0}  \\
\midrule
\textbf{\sname} & \textbf{50.5} & \textbf{73.5} & \underline{65.6}  & \textbf{43.5} & \textbf{1.3} \\
\bottomrule
\end{tabular}
}
\caption{Comparison of the performance of language models fine-tuned on GSM8k, evaluated on 11 tasks spanning four categories. Each category score is the average performance across tasks within the category, and the rank is computed as the average of the ranks within each category.}
\label{tab:gsm8k}
\end{table}

\begin{table}[t]
\centering
\resizebox{1.0\linewidth}{!}{%
\begin{tabular}{l|cccc|c}
\multicolumn{1}{c|}{} & \textbf{Math (3)} & \textbf{Knowl. (4)} & \textbf{Reason. (2)} & \textbf{Helpful. (2)} & \textbf{Rank} \\
\midrule
\rowcolor[rgb]{.816, .816, .816} \multicolumn{6}{l}{\textbf{Llama-3.1-8B}} \\
\midrule
Vanilla & 37.9 & 72.9 & 66.1 & 44.0 & -- \\
\midrule
SFT & 37.0    & 72.8  & 64.8  & \underline{50.6}  & 4.3 \\
NEFTune  & \underline{39.0}    & 72.6  & 64.2  & \underline{50.6}  & 4.3 \\
GEM   & 37.8  & 72.8  & \underline{65.1}  & \textbf{50.8} & 2.8 \\
IM    & 38.4  & \textbf{73.3} & \textbf{66.9} & 48.6  & \underline{2.5} \\
SDFT & 37.5  & \underline{73.2}  & 64.7  & 46.0     & 4.8 \\
\midrule
\textbf{\sname}  & \textbf{43.2} & \textbf{73.3} & 64.9  & \textbf{50.8} & \textbf{1.5} \\
\midrule
\rowcolor[rgb]{.816, .816, .816} \multicolumn{6}{l}{\textbf{Llama-3.1-8B-Instruct}} \\
\midrule
Vanilla & 60.5 & 74.0 & 81.0 & 69.9 & -- \\
\midrule
SFT & 56.6  & 72.9  & \underline{81.8}  & 69.3  & 4.8 \\
NEFTune   & 57.7  & 72.8  & \underline{81.8}  & \textbf{69.5} & 4.0  \\
GEM  & 59.0     & \underline{73.2}  & \textbf{81.9} & 68.7  & 3.5 \\
IM  & 59.6  & \textbf{73.6}  & \textbf{81.9} & 68.8  & \underline{2.5} \\
SDFT  & \underline{59.9}  & \underline{73.2}  & 81.2  & \underline{69.4}  & 3.5 \\
\midrule
\textbf{\sname} & \textbf{60.3} & \textbf{73.6}  & \textbf{81.9} & \textbf{69.5} & \textbf{1.0}  \\
\midrule
\rowcolor[rgb]{.816, .816, .816} \multicolumn{6}{l}{\textbf{Llama-2-7B}} \\
\midrule
Vanilla & 17.1 & 66.4 & 60.4 & 41.0 & --\\
\midrule
SFT   & 24.0    & 66.9  & 57.3  & \textbf{46.1}  & 3.3 \\
NEFTune & \underline{24.3}  & \textbf{67.1}  & 56.6  & 45.4  & 3.3 \\
GEM   & \underline{24.3}  & \underline{67.0}    & 57.3  & \underline{45.8}  & \underline{2.8} \\
IM    & 18.5  & 66.4  &\textbf{62.0}   & 41.2  & 4.3 \\
SDFT  & 12.8  & 66.1  & \underline{60.0}    & 43.0    & 4.8 \\
\midrule
\textbf{\sname} & \textbf{24.5}  & \underline{67.0}    & 57.8  & \textbf{46.1}  & \textbf{1.8} \\
\midrule
\rowcolor[rgb]{.816, .816, .816} \multicolumn{6}{l}{\textbf{Mistral-7B-v0.3}} \\
\midrule
Vanilla & 14.7 & 72.9 & 64.9 & 42.5 & -- \\
\midrule
SFT    & 28.9  & 72.9  & 65.5  & 45.4  & 4.5 \\
NEFTune & \underline{32.2}  & 73.0    & 65.4  & 47.9  & 3.8 \\
GEM   & 18.5  &\textbf{77.5}  & \underline{66.1}  & 44.6  & 3.5 \\
IM    & 23.4  & 73.1  &\textbf{67.1}  & 44.6  & \underline{3.3} \\
SDFT  & 22.3  & 73.0    & 65.8  & \underline{48.5}  & 3.5 \\
\midrule
\textbf{\sname} & \textbf{34.9}  & \underline{73.2}  & 65.6  & \textbf{48.6}  & \textbf{2.0} \\
\bottomrule
\end{tabular}

}
\caption{Comparison of the performance of language models fine-tuned on Alpagasus Dolly 3k, evaluated on 11 tasks spanning four categories. Each category score is the average performance across tasks within the category, and the rank is computed as the average of the ranks within each category.}
\label{tab:dolly}
\end{table}
\section{Experiments}

To evaluate the effectiveness of \sname, we conduct extensive experiments covering various adaptation scenarios of language models, including generalization to diverse task categories, applicability to various models, resilience to prompt variations and robustness against adversarial fine-tuning. We compare \sname with baselines \cite{jain2023neftune,li2025preserving,shi2024instruction,yang2024self} that aim to mitigate the overfitting issues of SFT. Detailed descriptions of these baselines are provided in Appendix~\ref{ap:baselines}. We begin by introducing the datasets used for evaluation.

\subsection{Setup}
\paragraph{Datasets}

{To comprehensively evaluate \sname, we adopt diverse datasets for fine-tuning and evaluation. For fine-tuning, we use the multi-task datasets Alpagasus Dolly 3k \cite{DatabricksBlog2023DollyV2} and LIMA \cite{zhou2023lima}, and the single-task datasets GSM8k \cite{cobbe2021training} and ARC-Challenge \cite{Clark2018ThinkYH}.
For evaluation, we cover four categories: \textbf{(\textit{i})} Math (GSM8k \cite{cobbe2021training}; MathQA \cite{amini2019mathqa}; ASDiv \cite{miao2021diverse}), \textbf{(\textit{ii})} Knowledge (MMLU \cite{hendryckstest2021}; PIQA \cite{Bisk2020}; HellaSwag \cite{zellers2019hellaswag}; LAMBADA \cite{paperno2016lambada}), \textbf{(\textit{iii})} Reasoning (ARC \cite{Clark2018ThinkYH}; CoQA \cite{reddy2019coqa}), and \textbf{(\textit{iv})} Helpfulness (ToxiGen \cite{hartvigsen2022toxigen}; TruthfulQA \cite{lin-etal-2022-truthfulqa}). Each category score is the average performance across tasks within the category, and the rank is computed as the average of the ranks across categories. Further dataset details are provided in Appendix~\ref{ap:dataset}.

\begin{figure}[t]
    \centering
    \includegraphics[width=0.85\linewidth]{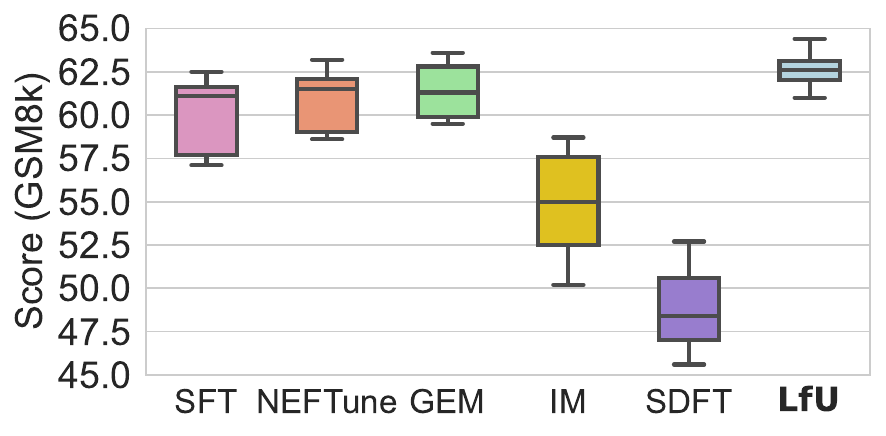}
    \caption{Performance distribution of GSM8k-fine-tuned Llama-3.1-8B models on five prompt variations of GSM8k. We use ChatGPT for generating the variations. \sname achieves the highest average accuracy and the lowest standard deviation, demonstrating strong resilience to prompt variations.}
    \label{fig:prompt_ago}
\end{figure}

\subsection{Results}

\paragraph {Single-task fine-tuning}

To demonstrate that \sname generalizes across diverse task categories and is applicable to various models, we evaluate it on four different language models: Llama-3.1-8B, Llama-3.1-8B-Instruct, Llama-2-7B, and Mistral-7B-v0.3.
These models are fine-tuned on GSM8k and evaluated on 11 tasks covering math, knowledge, reasoning, and helpfulness. As shown in Table~\ref{tab:gsm8k}, \sname achieves the best performance in all task categories on Llama-3.1-8B and records the highest rank. In particular, it improves performance by $+5.2\%$ in math and $+3.4\%$ in helpfulness compared to SFT. For Llama-3.1-8B-Instruct, \sname again achieves the highest rank. While SFT performs worse than the vanilla model in reasoning, \sname surpasses the vanilla model in this category. In the case of Llama-2-7B, \sname also obtains the highest rank among all baselines. In contrast, SFT underperforms the vanilla model on all tasks except math, whereas \sname improves performance on both math and knowledge. Mistral-7B-v0.3 shows a similar trend, with \sname achieving the highest rank. While SFT degrades performance in knowledge and reasoning, \sname enhances results across all task categories, including a $+4.8\%$ improvement in math. {\sname also shows a similar trend on ARC-Challenge and performs best on Qwen3 \cite{qwen3technicalreport} when fine-tuned on GSM8k. Detailed results are provided in Appendix~\ref{ap:additional_results}.}
These results demonstrate that \sname is effective across various models and generalizes well to diverse tasks.

\paragraph {Multi-task fine-tuning}
We next evaluate \sname in a multi-task setting using Alpagasus Dolly 3k and LIMA. We first present results from training with Alpagasus Dolly 3k.
According to the results in Table~\ref{tab:dolly}, \sname also demonstrates strong performance in the multi-task setting. In the case of LLaMA-3.1-8B, \sname achieves the highest rank, with a particularly notable improvement of $+16.8\%$ on the math tasks compared to SFT, which even underperforms the vanilla model.
For LLaMA-3.1-8B-Instruct, \sname achieves the best performance across all categories and improves the math category score by $+6.5\%$ over SFT. In LLaMA-2-7B, \sname also records the highest rank among all methods. On Mistral-7B-v0.3, \sname improves performance by $+20.8\%$ on math and $+7.0\%$ on helpfulness compared to SFT, once again achieving the highest rank. These results demonstrate that \sname is effective not only across a wide range of models and task categories, but also in both single-task and multi-task adaptation scenarios. \sname also shows strong performance when fine-tuned with LIMA, and the corresponding results are reported in Appendix~\ref{ap:additional_results}.

\paragraph{Representation steering (RepS) based LfU}
\begin{table}[t]
\centering
\resizebox{1.0\linewidth}{!}{%
\begin{tabular}{l|cccc|c}
\multicolumn{1}{c|}{} & \textbf{Math} & \textbf{Knowl.} & \textbf{Reason.} & \textbf{Helpful.}  & \textbf{Time/step (ms)} \\
\midrule
Vanilla & 37.9 & 72.9 & 66.1 & 44.0 & -- \\
\midrule
SFT & 51.5  & \underline{73.3}  & \underline{65.1} & 44.1  & {1386.9} $\pm$ 6.4  \\
\midrule
\textbf{\sname} (LoRA) & \textbf{54.2} & \textbf{73.5} & \textbf{65.6} & \underline{45.7}  &  4142.4 $\pm$ 14.2\\
\textbf{\sname} (RepS) & \underline{53.2} & \underline{73.3} &  \textbf{65.6}& \textbf{48.5}  &  2332.1 $\pm$ 9.1\\
\bottomrule
\end{tabular}

}
\caption{Comparison of the performance of Llama-3.1-8B fine-tuned on GSM8k and computation cost measured on a single NVIDIA H100 (80GB) instance, evaluated on 11 tasks spanning four categories: Math (3 tasks), Knowledge (4 tasks), Reasoning (2 tasks), and Helpfulness (2 tasks).}
\label{tab:abl_comp}
\end{table}

By default, we consider the LoRA-based design in our experiments for running LfU. In Table~\ref{tab:abl_comp}, we test the representation steering (RepS) based design in comparison to the LoRA version, particulary focusing on its effect in accelerating the LfU training. Specifically, we fine-tune Llama-3.1-8B on GSM8k using the two designs of LfU, and compare their performances across 11 comprehensive tasks, as well as their per-step computation time measured in a single NVIDIA H100 (80GB) instance. The results show that \sname (RepS) is nearly twice as fast as \sname (LoRA), while achieving competitive performance with the LoRA version but with a slight decrease in the in-domain performance.

\paragraph{Robustness to prompt variations}

When evaluating LMs on various tasks, it is known that performance can be highly sensitive to prompts \cite{reynolds2021prompt, zhu2023promptbench}.
To examine whether \sname is less sensitive to such prompt variation,
We evaluate GSM8k-adapted models on the GSM8k using five different prompt variations generated by ChatGPT \cite{openai2024gpt4ocard}, with examples listed in Appendix~\ref{ap:prompt_var}. As shown in Figure~\ref{fig:prompt_ago}, \sname consistently yields a $92.1\%$ smaller standard deviation and consistently outperforms other baselines. This indicates that \sname demonstrates resilience to prompt variations.

\paragraph{Robustness to adversarial fine-tuning}
\label{sec:bad-setup}
\begin{table}[t]
\centering
\resizebox{1.0\linewidth}{!}{%
\begin{tabular}{l|cc|cc|cc}
\multicolumn{1}{c|}{} & \multicolumn{2}{c|}{\textbf{HEx-PHI}} & \multicolumn{2}{c|}
{\textbf{PureBad}} & \multicolumn{2}{c}{\textbf{AdvBench}} \\
\midrule
Method & 0-step & 5-step & 0-step & 5-step & 0-step & 5-step \\
\midrule
SFT     & \underline{2.7}  & 34.8 & \textbf{3.0}   & 55.0   & \underline{0.2} & 7.3 \\
NEFTune & \textbf{2.4}  & 29.1 & \textbf{3.0}   & \underline{37.0}   & \textbf{0.0}   & \underline{4.8} \\
GEM     & 3.0  & 41.2 & \underline{4.0}   & 60.0   & \textbf{0.0}   & 21.7 \\
IM      & 5.8  & 60.9 & 8.0   & 80.0   & 1.2 & 81.9 \\
SDFT    & 6.1  & \underline{24.2} & 12.0  & 45.0   & 5.4 & 10.8 \\
\midrule
\textbf{LfU} & \textbf{2.4}  & \textbf{5.8} & \underline{4.0}   & \textbf{10.0} & \textbf{0.0}   & \textbf{0.6} \\
\bottomrule
\end{tabular}
}
\caption{Attack Success Rate (ASR) on HEx-PHI, PureBad and Advbench after 5 steps of adversarial fine-tuning on BeaverTails. We first align a Llama-3.1-8B via each baseline (or \sname) with the harmless subset of BeaverTails, and then continue fine-tuning the model on the harmful subset (of BeaverTails) using SFT.
}

\label{tab:adv}
\end{table}

{To empirically demonstrate the vulnerability of safety-aligned models to a few steps of adversarial fine-tuning, we construct safe BeaverTails \cite{ji2023beavertails} with 5,000 refusal instruction–output pairs, while harmful BeaverTails contains separate instructions paired with harmful outputs \cite{huang2024booster}. We first safety-align Llama-3.1-8B using the safe BeaverTails for 3 epochs using each baseline, then apply only 5 steps of adversarial fine-tuning with the harmful BeaverTails via SFT and measure attack success rate (ASR) on PureBad \cite{qi2023fine}, HEx-PHI \cite{qi2024safety}, and AdvBench \cite{zou2023universal}. As shown in Table~\ref{tab:adv}, \sname consistently demonstrates the highest robustness after adversarial training. Notably, compared to the second most robust method, \sname reduces ASR by $18.4\%$ on HEx-PHI and $27.0\%$ on PureBad. Furthermore, it achieves near-zero ASR on AdvBench, highlighting its strong robustness against a few steps of adversarial fine-tuning.}

\paragraph{Robustness to input noise}
\begin{figure}[t]  
\centering
\includegraphics[width=\linewidth]{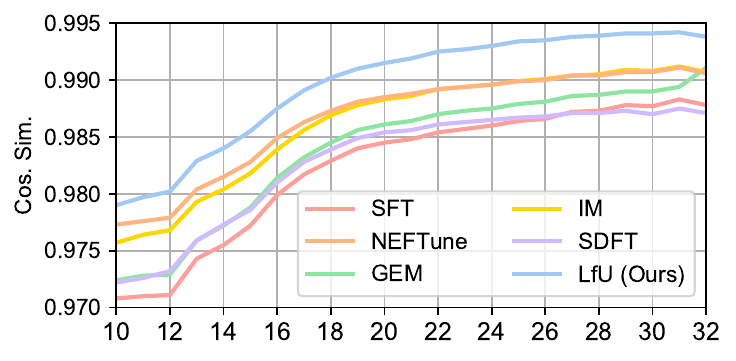}
\caption{Cosine similarity of internal representations between clean and noisy inputs, using Llama-3.1-8B fine-tuned on GSM8k. Clean inputs correspond to the original training data, while noisy inputs are generated by adding Gaussian noise to their input embeddings. Among all methods, \sname shows the highest similarity across layers.}
\label{fig:sft_noise}
\end{figure}
{To evaluate resilience to input noise, we measure cosine similarity between internal representations of clean and noisy inputs across all layers using GSM8k samples. We fine-tune Llama-3.1-8B on GSM8k and add noise to the input embeddings with the same strength as in NEFTune. As shown in Figure~\ref{fig:sft_noise}, \sname maintains higher cosine similarity across layers than all baselines, even surpassing NEFTune despite using the same noise level. This suggests that leveraging corrupted representations offers greater internal stability than using noise.}

\subsection{Ablation study}

\begin{table}[t]
\centering
\resizebox{1.0\linewidth}{!}{%
\begin{tabular}{rrc|cccc}
 \textbf{$\lambda$}  & \textbf{$\alpha$} & \textbf{Layer}&\textbf{Math (3)} & \textbf{Knowl. (4)} & \textbf{Reason. (2)} & \textbf{Helpful. (2)} \\
 \midrule
\rowcolor[rgb]{.816, .816, .816}
5.0        & 0.1  &All     & 54.2 & 73.5& 65.6 & 45.7\\
0.0         & - & -   & 51.5 & 73.3  &  65.1& 44.1\\
\midrule
\multicolumn{7}{l}{\textbf{(a) Varying hyperparameters}} \\
\midrule
0.1 &\textcolor{lightgray} {0.1}  & \textcolor{lightgray}{All}  & 48.1 & 73.2 &65.4 & 44.2  \\
1.0   & \textcolor{lightgray}{0.1}  & \textcolor{lightgray}{All}  & 48.1 & 73.4 & 65.4 & 44.2 \\
 10.0      &\textcolor{lightgray} {0.1}  &\textcolor{lightgray}{All}          & 53.4 & 73.0   &  65.0     &  44.4   \\
\midrule
\textcolor{lightgray}{5.0}   &0.001  &\textcolor{lightgray} {All}    & 52.5 & 73.1 & 65.1 & 44.0 \\
\textcolor{lightgray}{5.0}   & 0.01 &\textcolor{lightgray} {All}          & 53.0 & 73.1 & 65.2 & 44.3 \\
\textcolor{lightgray}{5.0}     &0.5  &\textcolor{lightgray}{All}       &  44.5 & 60.4 & 54.1 & 43.4  \\
\midrule
\multicolumn{7}{l}{\textbf{(b) w/ Layer selection}} \\
\midrule
\textcolor{lightgray}{5.0}  &\textcolor{lightgray} {0.1}  &   Early        & 53.6 & 73.3	  &  65.5   & 44.4  \\
 \textcolor{lightgray}{5.0}     & \textcolor{lightgray}{0.1}  &  Middle    & 53.3 &73.3	  & 65.6	    & 44.1  \\
\textcolor{lightgray} {5.0}     &\textcolor{lightgray}{ 0.1}  & Late      & 53.6 & 72.6 &  64.3   &  45.7  \\
\bottomrule
\end{tabular}

}
\caption{Ablation study on Llama-3.1-8B fine-tuned on GSM8k, evaluated across 11 tasks spanning four categories.}
\label{tab:ablation}
\end{table}

In Table~\ref{tab:ablation}, we conduct an ablation study to better understand the impact of representation consistency. Specifically, we investigate (a) how varying hyperparameters affect performance and (b) how different layer selections influence the effectiveness of consistency regularization.

\paragraph{Effect of hyperparameters}
We first analyze the individual effects of hyperparameters of LfU in Table~\ref{tab:ablation}(a). As $\lambda$ increases, we observe an improvement in in-domain performance, \ie, Math, while the performance on out-of-domain tasks slightly decreases. This suggests that there is a trade-off between in-domain alignment and generalization, and the current choice of $\lambda=5.0$ yields a balance.
Regarding $\alpha$, we observe that an excessive local update, \eg, $\alpha = 0.5$, can severely distort the model, making the resulting undesirable representations less meaningful. However, as long as the model is not overly distorted, such updates can still improve performance in out-of-domain tasks. More results across $\alpha$ and $\lambda$ values can be found in Appendix~\ref{ap:ablation}.

\paragraph{Layer selection}
We also investigate several specific layer selections when computing the proposed consistency loss in Table~\ref{tab:ablation}(b), beyond our default design of choosing all layers. We group the Llama-3.1-8B layers into Early (1–11), Middle (12–22), and Late (23–32) stages, and apply the consistency loss to each group individually. All settings provide gains and applying the regularization across all layers yields the best performance. This implies that representation-level regularization benefits from being distributed throughout the entire model, rather than focusing only on a subset of layers.

\begin{table}[t]
\centering
\resizebox{1.0\linewidth}{!}{%
\begin{tabular}{>{\raggedright\arraybackslash}p{6cm}cccc}
\toprule
\textbf{Method} & \textbf{Math} & \textbf{Knowl.} & \textbf{Reason.} & \textbf{Helpful.} \\
\midrule
(a) $\ell_{\text{SFT}}(\vtheta)$ & 51.5 & \underline{73.3}  &  65.1& 44.1\\
\midrule
(b) $\ell_{\text{SFT}}\left(\vtheta + \alpha \cdot \tfrac{\nabla_{\vtheta} \ell_{\text{SFT}}}{\left\| \nabla_{\vtheta} \ell_{\text{SFT}} \right\|_2}\right)$ &  53.3  & \underline{73.3}  & \underline{65.5}  & 44.0  \\
(c) $\ell_{\text{SFT}}(\vtheta) + \lambda \cdot \sum_{t=1}^{T} \sum_{l=1}^{K} \|\vh_{l,t}\|_{1}$& 53.1 &72.2 & 63.6 & 43.4 \\
(d) $\ell_{\text{SFT}}(\vtheta) +\lambda \cdot  \sum_{t=1}^{T} \sum_{l=1}^{K} \|\vh_{l,t}\|_{2}$ & 53.6 &73.0 & \underline{65.5} & 43.3 \\
(e) $\ell_{\text{SFT}}(\vtheta)+\lambda \cdot \ell_{\text{logit.}}(\vtheta,\vtheta_{\text{aux}})$ & \underline{53.9} & 73.2 & 65.2 & \underline{44.9} \\
\midrule
\rowcolor[rgb]{.816, .816, .816}
(f) $\ell_{\text{SFT}}(\vtheta)+\lambda \cdot \ell_{\text{cons.}}(\vtheta,\vtheta_{\text{aux}})$ & \textbf{54.2} & \textbf{73.5} & \textbf{65.6} & \textbf{45.7} \\
\bottomrule
\end{tabular}
}
\caption{Variation of update dynamics: Llama-3.1-8B fine-tuned on GSM8k, evaluated across 11 tasks spanning four categories: Math (3 tasks), Knowledge (4 tasks), Reasoning (2 tasks), and Helpfulness (2 tasks).}
\label{tab:update_dynamics}
\end{table}
\paragraph{Alternative loss designs}
To further demonstrate the effectiveness of representation-level consistency, we compare \sname with alternative loss designs: (a) the vanilla SFT, (b) directly updating $\vtheta$ for local updates, which is essentially equivalent to SAM~\cite{foret2020sharpness}, {(c) feature-level L1 consistency, (d) feature-level L2 consistency} and (e) logit-level consistency 
instead of the representation-level design of LfU, defined as follows:
\begin{multline}
\ell_{\text{logit.}}(\vtheta,\vtheta_{\text{aux}})
\\
= \mathbb{E}_{(\vx, \vy) \sim \mathcal{D}} \left[
\frac{1}{T} \sum_{t=1}^{T}
\mathrm{KL} \left(
\sigma\left(\mathrm{detach}(\vz_t)\right)
\;\middle\|\;
\sigma(\vz_t')
\right)
\right],
\end{multline}
where $\vz_t$ and $\vz'_t$ denote the logits from the original and undesirable models, respectively, $\sigma$ denotes the softmax function, and $\mathrm{KL}$ is the Kullback–Leibler divergence.

The results are shown in Table~\ref{tab:update_dynamics}. We observe that enforcing consistency at representation-level clearly obtains better generalization than directly optimizing the parameters, {and simple L1/L2 penalties control only feature magnitudes, failing to preserve the semantic properties of representations.}
The logit-level consistency also have gains over SFT; but our design consistently outperforms it, suggesting that consistency in internal representations plays a more critical role than consistency in logits. More results on these variations, as well as comparisons with parameter-level averaging approaches including Exponential Moving Average (EMA), are provided in Appendix~\ref{ap:ablation}, where our representation-level consistency still yields the best performance.

\section{Conclusion}
\label{conclusion}

In this work, we introduced Learning-from-the-Undesirable (\sname), a method designed to mitigate overfitting in language models by stabilizing internal representations. By simulating undesirable updates and enforcing representation-level consistency, \sname improves generalization by reducing overfitting. Extensive experiments show that \sname consistently outperforms prior fine-tuning methods on in-domain tasks, out-of-domain generalization, resilience to prompt variations, and robustness against adversarial fine-tuning. Furthermore, \sname is broadly compatible with various models and performs well in both single-task and multi-task fine-tuning settings. To reduce computational overhead, we also introduce a lightweight variant based on representation steering.
Overall, \sname offers a practical and broadly applicable solution for adapting language models to downstream tasks while preserving their general capabilities.

\section*{Acknowledgments}
\label{acknowledgements}
This work was supported by 
the Institute of Information \& Communications Technology Planning \& Evaluation (IITP) grants funded by the Korea government (MSIT)
(No.~RS-2019-II190079, Artificial Intelligence Graduate School Program (Korea University); 
No.~IITP-2025-RS-2025-02304828, Artificial Intelligence Star Fellowship Support Program to Nurture the Best Talents; 
No.~IITP-2025-RS-2024-00436857, Information Technology Research Center (ITRC)), 
and the Korea Creative Content Agency grant funded by the Ministry of Culture, Sports and Tourism (No.~RS-2025-00345025).

\bibliography{aaai2026}

@misc{foret2020sharpness,
  title={Sharpness-aware minimization for efficiently improving generalization},
  author={Foret, Pierre and Kleiner, Ariel and Mobahi, Hossein and Neyshabur, Behnam},
  journal={arXiv preprint arXiv:2010.01412},
  year={2020}
}

@inproceedings{li2025preserving,
  title={Preserving diversity in supervised fine-tuning of large language models},
  author={Li, Ziniu and Chen, Congliang and Xu, Tian and Qin, Zeyu and Xiao, Jiancong and Luo, Zhi-Quan and Sun, Ruoyu},
  booktitle={ICLR},
  year={2025}
}

@misc{oikonomou2025sharpness,
  title={Sharpness-Aware Minimization: General Analysis and Improved Rates},
  author={Oikonomou, Dimitris and Loizou, Nicolas},
  journal={arXiv preprint arXiv:2503.02225},
  year={2025}
}

@misc{tahmasebi2024universal,
  title={A universal class of sharpness-aware minimization algorithms},
  author={Tahmasebi, Behrooz and Soleymani, Ashkan and Bahri, Dara and Jegelka, Stefanie and Jaillet, Patrick},
  journal={arXiv preprint arXiv:2406.03682},
  year={2024}
}

@misc{wei2019eda,
  title={Eda: Easy data augmentation techniques for boosting performance on text classification tasks},
  author={Wei, Jason and Zou, Kai},
  journal={arXiv preprint arXiv:1901.11196},
  year={2019}
}

@article{shumailov2024ai,
  title={AI models collapse when trained on recursively generated data},
  author={Shumailov, Ilia and Shumaylov, Zakhar and Zhao, Yiren and Papernot, Nicolas and Anderson, Ross and Gal, Yarin},
  journal={Nature},
  volume={631},
  number={8022},
  pages={755--759},
  year={2024},
  publisher={Nature Publishing Group UK London}
}

@misc{rogulsky2024effects,
  title={The Effects of Hallucinations in Synthetic Training Data for Relation Extraction},
  author={Rogulsky, Steven and Popovic, Nicholas and F{\"a}rber, Michael},
  journal={arXiv preprint arXiv:2410.08393},
  year={2024}
}

@misc{guo2023curious,
  title={The curious decline of linguistic diversity: Training language models on synthetic text},
  author={Guo, Yanzhu and Shang, Guokan and Vazirgiannis, Michalis and Clavel, Chlo{\'e}},
  journal={arXiv preprint arXiv:2311.09807},
  year={2023}
}

@article{fang2024bias,
  title={Bias of AI-generated content: an examination of news produced by large language models},
  author={Fang, Xiao and Che, Shangkun and Mao, Minjia and Zhang, Hongzhe and others},
  journal={Scientific Reports},
  volume={14},
  number={1},
  pages={5224},
  year={2024},
  publisher={Nature Publishing Group UK London}
}

@misc{jain2023neftune,
  title={Neftune: Noisy embeddings improve instruction finetuning},
  author={Jain, Neel and Chiang, Ping-yeh and Wen, Yuxin and Kirchenbauer, John and Chu, Hong-Min and Somepalli, Gowthami and Bartoldson, Brian R and Kailkhura, Bhavya and Schwarzschild, Avi and Saha, Aniruddha and others},
  journal={arXiv preprint arXiv:2310.05914},
  year={2023}
}

@misc{qi2024safety,
  title={Safety alignment should be made more than just a few tokens deep},
  author={Qi, Xiangyu and Panda, Ashwinee and Lyu, Kaifeng and Ma, Xiao and Roy, Subhrajit and Beirami, Ahmad and Mittal, Prateek and Henderson, Peter},
  journal={arXiv preprint arXiv:2406.05946},
  year={2024}
}

@article{shi2024instruction,
  title={Instruction tuning with loss over instructions},
  author={Shi, Zhengxiang and Yang, Adam and Wu, Bin and Aitchison, Laurence and Yilmaz, Emine and Lipani, Aldo},
  journal={NeurIPS},
  volume={37},
  pages={69176--69205},
  year={2024}
}

@misc{yadav2023symnoise,
  title={SymNoise: Advancing Language Model Fine-tuning with Symmetric Noise},
  author={Yadav, Abhay Kumar and Singh, Arjun},
  journal={arXiv preprint arXiv:2312.01523},
  year={2023}
}

@misc{touvron2023llama,
  title={Llama 2: Open foundation and fine-tuned chat models},
  author={Touvron, Hugo and Martin, Louis and Stone, Kevin and Albert, Peter and Almahairi, Amjad and Babaei, Yasmine and Bashlykov, Nikolay and others},
  journal={arXiv preprint arXiv:2307.09288},
  year={2023}
}

@misc{mann2020language,
  title={Language models are few-shot learners},
  author={Mann, Ben and Ryder, N and Subbiah, M and Kaplan, J and Dhariwal, P and Neelakantan, A and Shyam, P and Sastry, G and Askell, A and Agarwal, S and others},
  journal={arXiv preprint arXiv:2005.14165},
  volume={1},
  pages={3},
  year={2020}
}

@misc{chatgpt,
  author = {OpenAI},
  title = {{Introducing ChatGPT}},
  howpublished = "\url{https://openai.com/blog/chatgpt}",
  year = {2022}
}

@misc{openai2023gpt4,
      title={GPT-4 Technical Report}, 
      author={OpenAI},
      year={2023},
      eprint={2303.08774},
      archivePrefix={arXiv},
      primaryClass={cs.CL}
}

@misc{claude,
  author = {Anthropic},
  title = {{Introducing Claude}},
  howpublished = "\url{https://www.anthropic.com/index/introducing-claude}",
  year = {2023}
}

@misc{geminiteam2023gemini,
      title={Gemini: A Family of Highly Capable Multimodal Models}, 
      author={{Gemini Team}},
    journal={arXiv preprint arXiv:2312.11805},
   year={2023}
}

@article{hendryckstest2021,
  title={Measuring Massive Multitask Language Understanding},
  author={Dan Hendrycks and Collin Burns and Steven Basart and Andy Zou and Mantas Mazeika and Dawn Song and Jacob Steinhardt},
  journal={ICLR},
  year={2021}
}

@misc{cobbe2021training,
      title={Training Verifiers to Solve Math Word Problems},
      author={Karl Cobbe and Vineet Kosaraju and Mohammad Bavarian and Jacob Hilton and Reiichiro Nakano and Christopher Hesse and John Schulman},
      year={2021},
      eprint={2110.14168},
      archivePrefix={arXiv},
      primaryClass={cs.LG}
}

@inproceedings{zellers2019hellaswag,
    title={HellaSwag: Can a Machine Really Finish Your Sentence?},
    author={Zellers, Rowan and Holtzman, Ari and Bisk, Yonatan and Farhadi, Ali and Choi, Yejin},
    booktitle ={ACL},
    year={2019}
}

@inproceedings{lin-etal-2022-truthfulqa,
    title = "{T}ruthful{QA}: Measuring How Models Mimic Human Falsehoods",
    author = "Lin, Stephanie  and
      Hilton, Jacob  and
      Evans, Owain",
    booktitle = "ACL",
    month = may,
    year = "2022",
    address = "Dublin, Ireland",
    url = "https://aclanthology.org/2022.acl-long.229",
    doi = "10.18653/v1/2022.acl-long.229",
    pages = "3214--3252",
}

@article{rafailov2023direct,
  title={Direct preference optimization: Your language model is secretly a reward model},
  author={Rafailov, Rafael and Sharma, Archit and Mitchell, Eric and Manning, Christopher D and Ermon, Stefano and Finn, Chelsea},
  journal={NeurIPS},
  volume={36},
  pages={53728--53741},
  year={2023}
}

@article{ouyang2022training,
  title={Training language models to follow instructions with human feedback},
  author={Ouyang, Long and Wu, Jeffrey and Jiang, Xu and Almeida, Diogo and Wainwright, Carroll and Mishkin, Pamela and Zhang, Chong and Agarwal, Sandhini and Slama, Katarina and Ray, Alex and others},
  journal={NeurIPS},
  volume={35},
  pages={27730--27744},
  year={2022}
}

@inproceedings{finn2017model,
  title={Model-agnostic meta-learning for fast adaptation of deep networks},
  author={Finn, Chelsea and Abbeel, Pieter and Levine, Sergey},
  booktitle={ICML},
  pages={1126--1135},
  year={2017},
  organization={PMLR}
}

@article{rajeswaran2019meta,
  title={Meta-learning with implicit gradients},
  author={Rajeswaran, Aravind and Finn, Chelsea and Kakade, Sham M and Levine, Sergey},
  journal={NeurIPS},
  volume={32},
  year={2019}
}

@misc{luo2023empirical,
  title={An empirical study of catastrophic forgetting in large language models during continual fine-tuning},
  author={Luo, Yun and Yang, Zhen and Meng, Fandong and Li, Yafu and Zhou, Jie and Zhang, Yue},
  journal={arXiv preprint arXiv:2308.08747},
  year={2023}
}

@misc{jiang2025unlocking,
  title={Unlocking the power of function vectors for characterizing and mitigating catastrophic forgetting in continual instruction tuning},
  author={Jiang, Gangwei and Jiang, Caigao and Li, Zhaoyi and Xue, Siqiao and Zhou, Jun and Song, Linqi and Lian, Defu and Wei, Yin},
  journal={arXiv preprint arXiv:2502.11019},
  year={2025}
}

@misc{qi2023fine,
  title={Fine-tuning aligned language models compromises safety, even when users do not intend to!},
  author={Qi, Xiangyu and Zeng, Yi and Xie, Tinghao and Chen, Pin-Yu and Jia, Ruoxi and Mittal, Prateek and Henderson, Peter},
  journal={arXiv preprint arXiv:2310.03693},
  year={2023}
}

@misc{gekhman2024does,
  title={Does fine-tuning LLMs on new knowledge encourage hallucinations?},
  author={Gekhman, Zorik and Yona, Gal and Aharoni, Roee and Eyal, Matan and Feder, Amir and Reichart, Roi and Herzig, Jonathan},
  journal={arXiv preprint arXiv:2405.05904},
  year={2024}
}

@misc{lin2024flame,
  title={Flame: Factuality-aware alignment for large language models},
  author={Lin, Sheng-Chieh and Gao, Luyu and Oguz, Barlas and Xiong, Wenhan and Lin, Jimmy and Yih, Wen-tau and Chen, Xilun},
  journal={arXiv preprint arXiv:2405.01525},
  year={2024}
}

@article{sajjadi2016regularization,
  title={Regularization with stochastic transformations and perturbations for deep semi-supervised learning},
  author={Sajjadi, Mehdi and Javanmardi, Mehran and Tasdizen, Tolga},
  journal={NeurIPS},
  volume={29},
  year={2016}
}

@misc{inan2023llama,
  title={Llama guard: Llm-based input-output safeguard for human-ai conversations},
  author={Inan, Hakan and Upasani, Kartikeya and Chi, Jianfeng and Rungta, Rashi and Iyer, Krithika and Mao, Yuning and Tontchev, Michael and Hu, Qing and Fuller, Brian and Testuggine, Davide and others},
  journal={arXiv preprint arXiv:2312.06674},
  year={2023}
}

@misc{yang2024self,
  title={Self-distillation bridges distribution gap in language model fine-tuning},
  author={Yang, Zhaorui and Pang, Tianyu and Feng, Haozhe and Wang, Han and Chen, Wei and Zhu, Minfeng and Liu, Qian},
  journal={arXiv preprint arXiv:2402.13669},
  year={2024}
}

@misc{roziere2023code,
  title={Code llama: Open foundation models for code},
  author={Roziere, Baptiste and Gehring, Jonas and Gloeckle, Fabian and Sootla, Sten and Gat, Itai and Tan, Xiaoqing Ellen and Adi, Yossi and Liu, Jingyu and Sauvestre, Romain and Remez, Tal and others},
  journal={arXiv preprint arXiv:2308.12950},
  year={2023}
}

@article{patil2024gorilla,
  title={Gorilla: Large language model connected with massive apis},
  author={Patil, Shishir G and Zhang, Tianjun and Wang, Xin and Gonzalez, Joseph E},
  journal={NeurIPS},
  volume={37},
  pages={126544--126565},
  year={2024}
}

@article{lewis2020retrieval,
  title={Retrieval-augmented generation for knowledge-intensive {NLP} tasks},
  author={Lewis, Patrick and Perez, Ethan and Piktus, Aleksandra and Petroni, Fabio and Karpukhin, Vladimir and others},
  journal={NeurIPS},
  volume={33},
  pages={9459--9474},
  year={2020}
}

@inproceedings{reynolds2021prompt,
  title={Prompt programming for large language models: Beyond the few-shot paradigm},
  author={Reynolds, Laria and McDonell, Kyle},
  booktitle={CHI},
  pages={1--7},
  year={2021}
}

@misc{zhu2023promptbench,
  title={Promptbench: Towards evaluating the robustness of large language models on adversarial prompts},
  author={Zhu, Kaijie and Wang, Jindong and Zhou, Jiaheng and Wang, Zichen and Chen, Hao and others},
  journal={arXiv e-prints},
  pages={arXiv--2306},
  year={2023}
}

@article{hu2022lora,
  title={Lora: Low-rank adaptation of large language models.},
  author={Hu, Edward J and Shen, Yelong and Wallis, Phillip and Allen-Zhu, Zeyuan and Li, Yuanzhi and Wang, Shean and Wang, Lu and Chen, Weizhu and others},
  journal={ICLR},
  volume={1},
  number={2},
  pages={3},
  year={2022}
}

@misc{qwen3technicalreport,
      title={Qwen3 Technical Report}, 
      author={{Qwen Team}},
      year={2025},
      primaryClass={cs.CL},
      url={https://arxiv.org/abs/2505.09388}, 
}

@misc{amini2019mathqa,
    title={MathQA: Towards Interpretable Math Word Problem Solving with Operation-Based Formalisms},
    author={Aida Amini and Saadia Gabriel and Peter Lin and Rik Koncel-Kedziorski and others},
    year={2019}
}

@misc{miao2021diverse,
    title={A Diverse Corpus for Evaluating and Developing English Math Word Problem Solvers},
    author={Shen-Yun Miao and Chao-Chun Liang and Keh-Yih Su},
    year={2021},
    eprint={2106.15772},
    archivePrefix={arXiv},
    primaryClass={cs.AI}
}

@inproceedings{Bisk2020,
    author = {Yonatan Bisk and Rowan Zellers and
            Ronan Le Bras and Jianfeng Gao
            and Yejin Choi},
    title = {PIQA: Reasoning about Physical Commonsense in
           Natural Language},
    booktitle = {AAAI},
    year = {2020},
}

@misc{paperno2016lambada,
  title={The LAMBADA dataset: Word prediction requiring a broad discourse context},
  author={Paperno, Denis and Kruszewski, Germ{\'a}n and Lazaridou, Angeliki and Pham, Quan Ngoc and Bernardi, Raffaella and others},
  journal={arXiv preprint arXiv:1606.06031},
  year={2016}
}

@misc{Clark2018ThinkYH,
  title={Think you have solved question answering? try arc, the ai2 reasoning challenge},
  author={Clark, Peter and Cowhey, Isaac and Etzioni, Oren and Khot, Tushar and Sabharwal, Ashish and Schoenick, Carissa and Tafjord, Oyvind},
  journal={arXiv preprint arXiv:1803.05457},
  year={2018}
}

@article{reddy2019coqa,
  title={Coqa: A conversational question answering challenge},
  author={Reddy, Siva and Chen, Danqi and Manning, Christopher D},
  journal={TACL},
  volume={7},
  pages={249--266},
  year={2019},
  publisher={MIT Press One Rogers Street, Cambridge, MA 02142-1209, USA journals-info~…}
}

@inproceedings{hartvigsen2022toxigen,
  title={ToxiGen: A Large-Scale Machine-Generated Dataset for Implicit and Adversarial Hate Speech Detection},
  author={Hartvigsen, Thomas and Gabriel, Saadia and Palangi, Hamid and Sap, Maarten and Ray, Dipankar and Kamar, Ece},
  booktitle={ACL},
  year={2022}
}

@article{zhou2023lima,
  title={Lima: Less is more for alignment},
  author={Zhou, Chunting and Liu, Pengfei and Xu, Puxin and Iyer, Srinivasan and Sun, Jiao and Mao, Yuning and Ma, Xuezhe and Efrat, Avia and Yu, Ping and Yu, Lili and others},
  journal={NeurIPS},
  volume={36},
  pages={55006--55021},
  year={2023}
}

@misc{DatabricksBlog2023DollyV2,
    author    = {Mike Conover and Matt Hayes and Ankit Mathur and Jianwei Xie and Jun Wan and Sam Shah and others},
    title     = {Free Dolly: Introducing the World's First Truly Open Instruction-Tuned {LLM}},
    year      = {2023},
    url       = {https://www.databricks.com/blog/2023/04/12/dolly-first-open-commercially-viable-instruction-tuned-llm},
    urldate   = {2023-06-30}
}

@misc{kotha2023understanding,
  title={Understanding catastrophic forgetting in language models via implicit inference},
  author={Kotha, Suhas and Springer, Jacob Mitchell and Raghunathan, Aditi},
  journal={arXiv preprint arXiv:2309.10105},
  year={2023}
}

@misc{huang2024booster,
  title={Booster: Tackling Harmful Fine-tuning for Large Language Models via Attenuating Harmful Perturbation},
  author={Huang, Tiansheng and Hu, Sihao and Ilhan, Fatih and Tekin, Selim Furkan and Liu, Ling},
  journal={arXiv preprint arXiv:2409.01586},
  year={2024}
}

@incollection{thrun1998learning,
  title={Learning to learn: Introduction and overview},
  author={Thrun, Sebastian and Pratt, Lorien},
  booktitle={Learning to learn},
  pages={3--17},
  year={1998},
  publisher={Springer}
}

@misc{li2017meta,
  title={Meta-sgd: Learning to learn quickly for few-shot learning},
  author={Li, Zhenguo and Zhou, Fengwei and Chen, Fei and Li, Hang},
  journal={arXiv preprint arXiv:1707.09835},
  year={2017}
}

@inproceedings{tang2023consistency,
  title={Consistency regularization for generalizable source-free domain adaptation},
  author={Tang, Longxiang and Li, Kai and He, Chunming and Zhang, Yulun and Li, Xiu},
  booktitle={ICCV workshop},
  pages={4323--4333},
  year={2023}
}

@article{xie2020unsupervised,
  title={Unsupervised data augmentation for consistency training},
  author={Xie, Qizhe and Dai, Zihang and Hovy, Eduard and Luong, Thang and Le, Quoc},
  journal={NeurIPS},
  volume={33},
  pages={6256--6268},
  year={2020}
}

@article{sohn2020fixmatch,
  title={Fixmatch: Simplifying semi-supervised learning with consistency and confidence},
  author={Sohn, Kihyuk and Berthelot, David and Carlini, Nicholas and Zhang, Zizhao and Zhang, Han and others},
  journal={NeurIPS},
  volume={33},
  pages={596--608},
  year={2020}
}

@inproceedings{ni2025noise,
  title={Noise Consistency Regularization for Improved Subject-Driven Image Synthesis},
  author={Ni, Yao and Wen, Song and Koniusz, Piotr and Cherian, Anoop},
  booktitle={CVPR Workshop},
  pages={3116--3126},
  year={2025}
}

@misc{ba2024fill,
  title={Fill in the gaps: Model calibration and generalization with synthetic data},
  author={Ba, Yang and Mancenido, Michelle V and Pan, Rong},
  journal={arXiv preprint arXiv:2410.10864},
  year={2024}
}

@inproceedings{wang2025debiased,
  title={Debiased distillation for consistency regularization},
  author={Wang, Lu and Xu, Liuchi and Yang, Xiong and Huang, Zhenhua and Cheng, Jun},
  booktitle={AAAI},
  volume={39},
  pages={7799--7807},
  year={2025}
}

@article{feder2023data,
  title={Data augmentations for improved (large) language model generalization},
  author={Feder, Amir and Wald, Yoav and Shi, Claudia and Saria, Suchi and Blei, David},
  journal={NeurIPS},
  volume={36},
  pages={70638--70653},
  year={2023}
}

@misc{chen2023alpagasus,
  title={Alpagasus: Training a better alpaca with fewer data},
  author={Chen, Lichang and Li, Shiyang and Yan, Jun and Wang, Hai and Gunaratna, Kalpa and Yadav, Vikas and Tang, Zheng and Srinivasan, Vijay and Zhou, Tianyi and Huang, Heng and others},
  journal={arXiv preprint arXiv:2307.08701},
  year={2023}
}

@article{ji2023beavertails,
  title={Beavertails: Towards improved safety alignment of llm via a human-preference dataset},
  author={Ji, Jiaming and Liu, Mickel and Dai, Josef and Pan, Xuehai and Zhang, Chi and Bian, Ce and Chen, Boyuan and Sun, Ruiyang and Wang, Yizhou and Yang, Yaodong},
  journal={NeurIPS},
  volume={36},
  pages={24678--24704},
  year={2023}
}

@misc{zou2023universal,
  title={Universal and transferable adversarial attacks on aligned language models},
  author={Zou, Andy and Wang, Zifan and Carlini, Nicholas and Nasr, Milad and Kolter, J Zico and Fredrikson, Matt},
  journal={arXiv preprint arXiv:2307.15043},
  year={2023}
}

@misc{openai2024gpt4ocard,
      title={G{PT}-4o System Card}, 
      author={OpenAI},
      year={2024},
      eprint={2410.21276},
      archivePrefix={arXiv},
      primaryClass={cs.CL},
      url={https://arxiv.org/abs/2410.21276}, 
}

@inproceedings{zhang2020Consistency,
title={Consistency Regularization for Generative Adversarial Networks},
author={Han Zhang and Zizhao Zhang and Augustus Odena and Honglak Lee},
booktitle={ICLR},
year={2020},
url={https://openreview.net/forum?id=S1lxKlSKPH}
}

@article{jeong2020consistency,
  title={Consistency regularization for certified robustness of smoothed classifiers},
  author={Jeong, Jongheon and Shin, Jinwoo},
  journal={NeurIPS},
  volume={33},
  pages={10558--10570},
  year={2020}
}

@inproceedings{zhang2019theoretically,
  title={Theoretically principled trade-off between robustness and accuracy},
  author={Zhang, Hongyang and Yu, Yaodong and Jiao, Jiantao and Xing, Eric and El Ghaoui, Laurent and Jordan, Michael},
  booktitle={ICML},
  pages={7472--7482},
  year={2019}
}

@inproceedings{tack2022consistency,
  title={Consistency regularization for adversarial robustness},
  author={Tack, Jihoon and Yu, Sihyun and Jeong, Jongheon and Kim, Minseon and Hwang, Sung Ju and Shin, Jinwoo},
  booktitle={AAAI},
  volume={36},
  pages={8414--8422},
  year={2022}
}

@misc{kirk2023understanding,
  title={Understanding the effects of RLHF on LLM generalisation and diversity},
  author={Kirk, Robert and Mediratta, Ishita and Nalmpantis, Christoforos and Luketina, Jelena and Hambro, Eric and others},
  journal={arXiv preprint arXiv:2310.06452},
  year={2023}
}

\appendix
\section {Experimental details}
\label{ap:experimental details}
All methods are trained using Low-Rank Adaptation (LoRA) \cite{hu2022lora} with a rank of $r = 8$, and we ensure consistent training hyperparameters across methods for fair comparison. Specifically, all models are fine-tuned for 3 epochs with a batch size of 16 across all datasets, using a learning rate of 1e-4 and a cosine learning rate scheduler, unless otherwise specified. An exception is adversarial fine-tuning, which is conducted with a higher learning rate of 3e-4. Method-specific parameters are adopted as described in their respective formulations. For \sname(LoRA), we set $\alpha = 0.1$ and $\lambda = 5$, while for the Llama-3.1-8B-Instruct model we use $\alpha = 0.001$ and $\lambda = 500$. For \sname(RepS), we set $\alpha = 0.1$ and $\lambda = 0.1$. All experiments are conducted using two NVIDIA H100 GPUs and two NVIDIA A40 GPUs.

\section{Baselines}
\label{ap:baselines}
We compare our method against several recent approaches designed to improve the generalization of SFT.
\begin{itemize}
    \item \textbf{NEFTune} \cite{jain2023neftune} injects uniform noise into input embeddings during training to prevent overfitting.
\end{itemize}
\begin{itemize}
    \item \textbf{GEM} \cite{li2025preserving} reduces the log-likelihood gap between real and model-generated tokens, encouraging diversity by considering alternative outputs during training.
\end{itemize}
\begin{itemize}
    \item  \textbf{IM} \cite{shi2024instruction} applies loss to both instruction and output tokens (excluding prompt templates) to reduce overfitting during fine-tuning.
\end{itemize}
\begin{itemize}
    \item  \textbf{SDFT} \cite{yang2024self} improves generalization by generating a distilled dataset using the pre-trained model itself. The model regenerates outputs based on the original input-output pairs, and is then fine-tuned on this self-distilled data to mitigate distribution shift.
\end{itemize}

\section{Datasets}
\label{ap:dataset}

\subsection{Fine-tuning data}
We utilize four datasets for fine-tuning, which can be categorized into two types based on their task coverage: single-task and multi-task datasets.

\subsubsection{Single-task}

\begin{itemize}
    \item \textbf{GSM8k}: A dataset focused on grade-school-level mathematical reasoning, where each example includes a math word problem and a step-by-step solution.
    \item \textbf{ARC-Challenge} (Table~\ref{tab:arc}): A science exam question-answering dataset that involves advanced reasoning. 
\end{itemize}

\subsubsection{Multi-task}

\begin{itemize}
    \item \textbf{Alpagasus Dolly 3k}: This dataset is generated via Alpagasus \cite{chen2023alpagasus}, with outputs revised using GPT-4o-mini \cite{openai2024gpt4ocard}.
    \item \textbf{LIMA} (Table~\ref{tab:lima}): A benchmark consisting of 1,000 high-quality, human-curated instruction–output pairs. 

\end{itemize}

\begin{algorithm}[t]
\begin{algorithmic}[1]
\Require Model parameters $\vtheta$, inputs $(\vx, \vy)$
\Statex \hspace{2.6em}Model has $K$ layers: $l = 1, \dots, K$
\Require Hyperparameters $\alpha$, $\lambda$, learning rate $\gamma$
\For{epoch $= 1$ to $N$}
    \State Generate auxiliary model: $$\vtheta_{\text{aux}} = [\vtheta_1 + \mathrm{LoRA}_1, \dots, \vtheta_K + \mathrm{LoRA}_K]$$
    \State Compute $\ell_{\text{SFT}}(\vtheta_{\text{aux}})$
    \State Perturb parameters: $$\vtheta_{\text{aux}} \gets \vtheta_{\text{aux}} + \alpha \cdot \frac{\nabla_{\vtheta_{\text{aux}}} \ell_{\text{SFT}}(\vtheta_{\text{aux}})}{\left\|\nabla_{\vtheta_{\text{aux}}} \ell_{\text{SFT}}(\vtheta_{\text{aux}})\right\|_2}$$

    \For{$l = 1$ to $K$}
    \State $h_{l,t} \gets M^{(l)}(\vx, \vy_{<t}; \vtheta)$
    \State $h_{l,t}' \gets M^{(l)}(\vx, \vy_{<t}; \vtheta, \vtheta_{\text{aux}})$
    \EndFor
    \State Compute consistency loss: $$\ell_{\text{cons.}} = \frac{1}{TK} \sum_{t=1}^{T} \sum_{l=1}^{K} \left\| \mathrm{detach}(\vh_{l,t}) - \vh'_{l,t} \right\|^2$$
    \State Compute LfU loss: $$\ell_{\text{LfU}}(\vtheta,\vtheta_{\text{aux}}) = \ell_{\text{SFT}}(\vtheta) + \lambda \cdot \ell_{\text{cons.}}(\vtheta,\vtheta_{\text{aux}})$$
    \State Update model: $\vtheta \gets \vtheta - \gamma \cdot \nabla_\vtheta \ell_{\text{LfU}}(\vtheta,\vtheta_{\text{aux}})$
\EndFor
\end{algorithmic}

\caption{LfU (LoRA)}
\label{alg:lfu_lora}
\end{algorithm}

\subsection{Evaluation data}
To comprehensively assess the capabilities of fine-tuned models, we evaluate their performance across four key categories: Math, Knowledge, Reasoning, and Helpfulness.

\subsubsection{Math}

\begin{itemize}
    \item \textbf{GSM8k}: This dataset features multi-step grade school math problems. We use final numeric answer extraction, and report exact match accuracy as the evaluation metric.
    \item \textbf{MathQA}: Comprising multiple-choice math questions, this dataset assesses mathematical reasoning. Evaluation is conducted on the test split using mean accuracy.
    \item \textbf{ASDiv}: Designed to capture diverse elementary-level math problems, this dataset is evaluated using the Chain-of-Thought prompting format with final numeric answer extraction. The metric used is exact match accuracy.
\end{itemize}

\subsubsection{Knowledge}

\begin{itemize}
    \item \textbf{MMLU}: Encompassing 57 multiple-choice tasks from STEM, humanities, social sciences, and more, this dataset is evaluated on the test split using mean accuracy.
    \item \textbf{PIQA}: Designed to assess physical commonsense reasoning, PIQA is evaluated on the test split with mean accuracy over multiple-choice questions.
    \item \textbf{HellaSwag}: A dataset for commonsense natural language inference, consisting of sentence completion tasks. Evaluated on the test set using mean accuracy.
    \item \textbf{LAMBADA}: A benchmark where the task is to predict the final word of a passage. We report mean accuracy on the test split.
\end{itemize}

\subsubsection{Reasoning}

\begin{itemize}
    \item \textbf{ARC}: We use the ARC-Challenge subset, targeting complex science exam questions in multiple-choice. Evaluation is based on the mean accuracy on the test split.
    \item \textbf{CoQA}: A conversational question answering task over text passages, requiring multi-turn understanding. Evaluation is based on the F1 score on the test set.
\end{itemize}

\subsubsection{Helpfulness}

\begin{itemize}
    \item \textbf{ToxiGen}: This dataset presents a binary classification task to determine whether a text is hateful. We compute mean accuracy as the evaluation metric.
    \item \textbf{TruthfulQA}: Designed to assess a model’s ability to resist imitating human falsehoods, TruthfulQA (mc2) includes multiple-choice questions with potentially multiple correct answers. Evaluated using mean accuracy.
\end{itemize}

\begin{algorithm}[!t]
\begin{algorithmic}[1]
\Require Model parameters $\vtheta$, inputs $(\vx, \vy)$, 
\Statex \hspace{2.6em}Steering vectors $\{\vd_l\}_{l=1}^K$,
\Statex \hspace{2.6em}Model has $K$ layers: $l = 1, \dots, K$
\Require Hyperparameters $\alpha$, $\lambda$, learning rate $\gamma$
\For{each epoch $= 1$ to $N$}
    \State Generate auxiliary model: 
    \begin{multline*}
    \vtheta_{\text{aux}} = \vtheta \text{ with } \vh_{l,t}' \leftarrow \vh_{l,t} + \vd_l, \\
    \text{for all } l \in [1, K],\; t \in [1, T]
    \end{multline*}
    \State Compute $\ell_{\text{SFT}}( \vtheta_{\text{aux}})$
    \State Perturb parameters: $$\vtheta_{\text{aux}} \gets \vtheta_{\text{aux}} + \alpha \cdot \frac{\nabla_{\vtheta_{\text{aux}}} \ell_{\text{SFT}}(\vtheta_{\text{aux}})}{\left\|\nabla_{\vtheta_{\text{aux}}} \ell_{\text{SFT}}(\vtheta_{\text{aux}})\right\|_2}$$
    \For{$l = 1$ to $K$}
        \State $h_{l,t} \gets M^{(l)}(\vx, \vy_{<t}; \vtheta)$
        \State $h_{l,t}' \gets M^{(l)}(\vx, \vy_{<t}; \vtheta, \vtheta_{\text{aux}})$
    \EndFor
    
    \State Compute consistency loss: 
    $$\ell_{\text{cons.}} = \frac{1}{TK} \sum_{t=1}^{T} \sum_{l=1}^{K} \left\| \mathrm{detach}(\vh_{l,t}) - \vh'_{l,t} \right\|^2$$
    \State Compute LfU loss: $$\ell_{\text{LfU}}(\vtheta,\vtheta_{\text{aux}}) = \ell_{\text{SFT}}(\vtheta) + \lambda \cdot \ell_{\text{cons.}}(\vtheta,\vtheta_{\text{aux}})$$ 
    \State Update model: $\vtheta \gets \vtheta - \gamma \cdot \nabla_\vtheta \ell_{\text{LfU}}(\vtheta,\vtheta_{\text{aux}})$
\EndFor
\end{algorithmic}

\caption{LfU (RepS)}
\label{alg:lfu_RepS}
\end{algorithm}

\subsection{Adversarial data}

To evaluate robustness to adversarial fine-tuning, we use four datasets. Following \citet{qi2023fine}, we assess the safety of models’ outputs through keyword matching. \\

\begin{itemize}
    \item \textbf{BeaverTails}: A dataset of question–answer pairs covering various harm types. Following \citet{huang2024booster}, we construct two versions: safe BeaverTails and harmful BeaverTails.
    \item \textbf{HEx-PHI}: Collection of 330 harmful instructions.
    \item \textbf{PureBad}: 100 harmful instructions.
    \item \textbf{AdvBench}: Dataset of 500 harmful instructions.
\end{itemize}

\begin{table}[t]
\centering
\resizebox{1.0\linewidth}{!}{%
\begin{tabular}{l|cccc|c}
\multicolumn{1}{c|}{} & \textbf{Math (3)} & \textbf{Knowl. (4)} & \textbf{Reason. (2)} & \textbf{Helpful. (2)}  & \textbf{Rank} \\
\midrule
\rowcolor[rgb]{.816, .816, .816} \multicolumn{6}{l}{\textbf{Llama-3.1-8B}} \\
\midrule
Vanilla & 37.9 & 72.9 & 66.1 & 44.0 & -- \\
\midrule
SFT & 36.7  & \underline{73.2}  & \textbf{80.8}  & 44.9  & 3.0 \\
NEFTune  & \underline{37.6}  & \textbf{73.3}  & 79.4  & \textbf{49.8}  & \underline{2.0} \\
GEM   &  \underline{37.6}  & \underline{73.2}  & \underline{80.5}  & 45.0    & 2.5 \\
IM   & 36.3  &\underline{73.2}  & 79.1  & 44.1  & 4.8 \\
SDFT  & 35.3  & 72.9  & 79.3  & 41.3  & 5.8 \\
\midrule
\textbf{\sname}&\textbf{37.9}  & \textbf{73.3}  & \underline{80.5}  & \underline{45.2}  & \textbf{1.5} \\
\midrule
\rowcolor[rgb]{.816, .816, .816} \multicolumn{6}{l}{\textbf{Llama-3.1-8B-Instruct}} \\
\midrule
Vanilla & 60.5 & 74.0 & 81.0 & 69.9 & -- \\
\midrule
SFT & 60.8  & \textbf{74.0}    & 81.4  & 70.1  & 3.5 \\
NEFTune &\underline{61.2}  & \underline{73.9}  & 81.6  & \underline{70.5}  & 2.8 \\
GEM  & 60.4  & \textbf{74.0}    & \textbf{82.2}  & \underline{70.5}  & \underline{2.3} \\
IM  & 61.1  & 73.4  & 81.9  & 68.0    & 4.0 \\
SDFT  &  60.0    & 73.3  & 79.5  & 67.3  & 6.0 \\
\midrule
\textbf{\sname} & \textbf{61.3}  & \underline{73.9}  & \underline{82.0}    & \textbf{70.6}  & \textbf{1.8} \\
\midrule
\rowcolor[rgb]{.816, .816, .816} \multicolumn{6}{l}{\textbf{Llama-2-7B}} \\
\midrule
Vanilla & 17.1  & 66.4  & 60.4  & 41.0     & -- \\
\midrule
SFT  & 24.7  & 68.2  & 69.5  & 39.9  & 3.5 \\
NEFTune  & \underline{25.0}    & 68.2  & 69.2  & 39.8  & 4.0 \\
GEM & 24.2  & \underline{68.3}  & \textbf{70.5}  & \underline{40.5}  & \underline{2.5} \\
IM & 24.4  & 67.5  & 69.3  & \underline{40.5}  & 3.8 \\
SDFT  &23.1  & 65.0    & 69.1  & \textbf{40.8}  & 4.8 \\
\midrule
\textbf{\sname} & \textbf{25.7}  & \textbf{68.6}  & \underline{70.1}  & \underline{40.5}  & \textbf{1.5} \\
\midrule
\rowcolor[rgb]{.816, .816, .816} \multicolumn{6}{l}{\textbf{Mistral-7B-v0.3}} \\
\midrule
Vanilla & 14.7  & 72.9  & 64.9  & 42.5  & -- \\
\midrule
SFT & 20.3  & 73.1  & 79.3  & 43.0    & 4.3 \\
NEFTune &  \textbf{22.4}  & 72.7  &\underline{79.4}  & 44.0    & 3.0 \\
GEM  & 21.0    & \textbf{73.5}  & \textbf{79.7}  & 42.8  & \underline{2.8} \\
IM  &  21.2  & \underline{73.3}  & 76.8  & \underline{44.6}  & 3.3 \\
SDFT  &20.9  & 73.1  & 79.0   & 42.7  & 5.0 \\
\midrule
\textbf{\sname} & \underline{21.3}  & 73.2  & 79.3  & \textbf{45.0}    & \textbf{2.3} \\
\bottomrule
\end{tabular}
}
\caption{Comparison of the performance of language models fine-tuned on ARC-Challenge, evaluated on 11 tasks spanning four categories. Each category score is the average performance across tasks within the category, and the rank is computed as the average of the ranks within each category.}
\label{tab:arc}
\end{table}

\subsection{Prompt variations}
\label{ap:prompt_var}
{
To assess robustness to prompt variations, we introduce several alternative prompt formulations for GSM8k. We leverage ChatGPT \cite{openai2024gpt4ocard} to generate these variations. The following examples illustrate the types of prompt templates used in our experiments.

\begin{itemize}
    \item \texttt{Question: \{question\}}
    \item \texttt{Problem: \{question\}}
    \item \texttt{Let's solve this: \{question\}}
    \item \raggedright\texttt{You are given the following: \{question\}}
    \item \texttt{Here’s the question: \{question\}}
\end{itemize}
The \texttt{{\{question\}}} is a sample from the GSM8k test set.}

\section {Overall procedure of LfU}
\label{ap:pseudo_lfu}
We provide the detailed pseudocode for LfU with a LoRA based auxiliary model (LfU (LoRA)) in Algorithm~\ref{alg:lfu_lora} and for LfU with a representation steering based auxiliary model (LfU (RepS)) in Algorithm~\ref{alg:lfu_RepS}.

\begin{table}[t]
\centering
\resizebox{1.0\linewidth}{!}{%
\begin{tabular}{l|cccc|c}
\multicolumn{1}{c|}{} & \textbf{Math (3)} & \textbf{Knowl. (4)} & \textbf{Reason. (2)} & \textbf{Helpful. (2)}  & \textbf{Rank} \\
\midrule
\rowcolor[rgb]{.816, .816, .816} \multicolumn{6}{l}{\textbf{Llama-3.1-8B}} \\
\midrule
Vanilla & 37.9 & 72.9 & 66.1 & 44.0 & -- \\
\midrule
SFT & 33.8  & 73.5  & 65.5  & 45.7  & 4.0 \\
NEFTune  & 33.8  & \underline{73.6}  & 65.7  & \underline{45.8}  & 2.8 \\
GEM   &  32.7  & 73.5  & 65.1  & \textbf{46.5}  & 4.0 \\
IM   & \textbf{36.4}  & \underline{73.6}  & \textbf{66.3}  & 45.5  & \underline{2.3} \\
SDFT  & 33.9  & 73.3  & 64.9  & 42.6  & 5.3 \\
\midrule
\textbf{\sname}& \underline{34.8}  &\textbf{73.8}  & \underline{65.9}  & \underline{45.8}  & \textbf{1.8} \\
\midrule
\rowcolor[rgb]{.816, .816, .816} \multicolumn{6}{l}{\textbf{Llama-3.1-8B-Instruct}} \\
\midrule
Vanilla & 60.5 & 74.0 & 81.0 & 69.9 & -- \\
\midrule
SFT & 52.4  & 73.7  & 80.0    & 64.3  & 4.5 \\
NEFTune &53.8  & 73.7  & 79.8  & 64.8  & 4.0 \\
GEM  & 52.9  & \textbf{73.9}  & 80.5  & 65.0    & 3.0 \\
IM  & 51.8  & \underline{73.8}  & \textbf{81.9}  & \textbf{66.2}  & \underline{2.5} \\
SDFT  &  \textbf{56.9}  & 73.3  & 80.6  & 58.8  & 4.0 \\
\midrule
\textbf{\sname} & \underline{55.1}  & 73.7  & \underline{80.7}  & \underline{66.1}  & \textbf{2.3} \\
\midrule
\rowcolor[rgb]{.816, .816, .816} \multicolumn{6}{l}{\textbf{Llama-2-7B}} \\
\midrule
Vanilla & 17.1  & 66.4  & 60.4  & 41.0     & -- \\
\midrule
SFT  & \textbf{17.4}  & \underline{66.5}  & 60.1  & 40.9  & 3.3 \\
NEFTune  & \underline{17.3}  & \textbf{66.6}  & 59.8  & 41.0    & \underline{3.0} \\
GEM & 17.1  & \underline{66.5}  & 60.1  & \textbf{43.4}  & \underline{3.0} \\
IM & 16.6  & 66.0    & \textbf{61.5}  & 40.5  & 4.8 \\
SDFT  & 17.0    & \textbf{66.6}  & 59.3  & 41.5  & 3.8 \\
\midrule
\textbf{\sname} & \underline{17.3}  & \textbf{66.6}  & \underline{60.4}  & \underline{41.7}  & \textbf{1.8} \\
\midrule
\rowcolor[rgb]{.816, .816, .816} \multicolumn{6}{l}{\textbf{Mistral-7B-v0.3}} \\
\midrule
Vanilla & 14.7  & 72.9  & 64.9  & 42.5  & -- \\
\midrule
SFT & \underline{32.7}  & \underline{73.7}  & \underline{65.5}  & 46.2  & 3.5 \\
NEFTune &  \textbf{33.4}  &\textbf{73.9}  & \underline{65.5}  & \underline{47.4}  & \underline{1.8} \\
GEM  & 32.1  & \underline{73.7}  & 65.2  & \textbf{47.5}  & 3.3 \\
IM  &  24.9  & 72.7  & \textbf{66.1}  & 45.2  & 4.8 \\
SDFT  &31.6  & 73.4  & 64.7  & 46.3  & 5.0 \\
\midrule
\textbf{\sname} & \textbf{33.4}  & \textbf{73.9}  & \textbf{66.1}  & 46.5  & \textbf{1.5} \\
\bottomrule
\end{tabular}

}
\caption{Comparison of the performance of language models fine-tuned on LIMA, evaluated on 11 tasks spanning four categories. Each category score is the average performance across tasks within the category, and the rank is computed as the average of the ranks within each category.}
\label{tab:lima}
\end{table}

\section{Additional results}
\label{ap:additional_results}

\paragraph{Single-task fine-tuning on ARC-Challenge}
We provide additional results on the single-task dataset ARC-Challenge in Table~\ref{tab:arc}, where \sname consistency ranks highest compared to other baselines. Achieving strong performance in both in-domain and out-of-domain categories demonstrates its ability to generalize effectively.

\paragraph{Multi-task fine-tuning on LIMA}
We further present the fine-tuning results on the multi-task dataset LIMA in Table~\ref{tab:lima}, where \sname ranks highest among Llama-3.1-8B, Llama-3.1-8B-Instruct, Llama-2-7B, and Mistral-7B-v0.3. Moreover, it achieves consistently higher performance than SFT across all categories for every model, demonstrating strong generalization across diverse models.

\begin{table}[t]
\centering
\resizebox{1.0\linewidth}{!}{%
\begin{tabular}{l|cccc|c}
\multicolumn{1}{c|}{} & \textbf{Math (3)} & \textbf{Knowl. (4)} & \textbf{Reason. (2)} & \textbf{Helpful. (2)}  & \textbf{Rank} \\
\midrule
Vanilla & 30.0  & 71.8  & 69.1  & 53.1  & -- \\
\midrule
SFT & 67.0    & 73.5  & 68.3  & 52.4  & 4.3 \\
NEFTune &  67.0    & 73.4  & 68.3  & 53.6  & 4.8 \\
GEM  & 67.3  & \underline{73.6}  & 67.9  & 54.8  & 3.5 \\
IM  &  66.3  & 73.5  & \textbf{69.6}  & \underline{55.0}    & \underline{3.0} \\
SDFT  &\underline{68.7}  & 73.5  & 68.5  & 54.0    & \underline{3.0} \\
\midrule
\textbf{\sname} & \textbf{68.8}  & \textbf{73.7}  & \underline{69.1}  & \textbf{55.4}  & \textbf{1.3} \\
\bottomrule
\end{tabular}
}
\caption{{Comparison of the performance of Qwen3 fine-tuned on GSM8k, evaluated on 11 tasks spanning four categories. Each category score is the average performance across tasks within the category, and the rank is computed as the average of the ranks within each category.}}
\label{tab:qwen3}
\end{table}

\begin{table}[!htbp]
\centering
\resizebox{1.0\linewidth}{!}{%
\begin{tabular}{l|c|c|c}
\multicolumn{1}{c|}{Time/step (ms)} & Llama-3.1-8B &  Llama-2-13B & Ratio (13B/8B)  \\
\midrule
    SFT & 1386.9 $\pm$ 6.4   & 1598.6 $\pm$ 8.3  & 1.15 \\
\midrule
    \textbf{\sname} (LoRA)  & 4142.4 $\pm$ 14.2   & 4790.9 $\pm$ 13.6  & 1.16   \\
    \textbf{\sname} (RepS)  & 2332.1 $\pm$ 9.1   & 2820.1 $\pm$ 9.7  & 1.21   \\
\bottomrule
\end{tabular}
}
\caption{{Comparison of computational costs for a single-task fine-tuning on GSM8K, measured on a single NVIDIA H100 (80 GB) instance.}}
\label{tab:ap_comp}
\end{table}

\begin{table}[!htbp]
\centering
\resizebox{1.0\linewidth}{!}{%
\begin{tabular}{>{\raggedright\arraybackslash}p{5.5cm}cccc}
\toprule
\textbf{Method} & \textbf{Math} & \textbf{Knowl.} & \textbf{Reason.} & \textbf{Helpful.} \\
\midrule
SFT & 51.5 & 73.3 &  65.1& 44.1\\
\midrule
(a) $\vtheta_{\text{ens.}} \!\leftarrow\!  \beta \cdot\,\vtheta_{\text{vanilla}} + (1-\beta)\cdot\vtheta_{\text{SFT}}$ & 53.1 & \underline{73.4}& \underline{65.4}& 44.3 \\
(b) $\vtheta_{\text{EMA}} \!\leftarrow\! \beta\cdot\vtheta_{\text{EMA}} + (1-\beta)\cdot\vtheta_{\text{current}}$ & \underline{53.4} & 71.6 & 61.3& \underline{44.4} \\
\midrule
\textbf{\sname} & \textbf{54.2} & \textbf{73.5} & \textbf{65.6} & \textbf{45.7} \\
\bottomrule
\end{tabular}
}
\caption{{Comparison with parameter-level averaging methods (Ensembling and EMA). All results are obtained on Llama-3.1-8B fine-tuned on GSM8k and evaluated across 11 tasks spanning four categories: Math (3 tasks), Knowledge (4 tasks), Reasoning (2 tasks), and Helpfulness (2 tasks).}}
\label{tab:parameter}
\end{table}

\begin{table}[!htbp]
\centering
\resizebox{1.0\linewidth}{!}{%
\begin{tabular}{r|c|c|c|c|c}
\multicolumn{1}{c|} {}&    Epoch 0.5 & Epoch 1&  Epoch 2 &  Epoch 3 &  Epoch 4 \\
\midrule
SFT   & \textbf{57.7}  & \textbf{60.2}  & 61.2 & 61.6 & 61.8 \\
\textbf{\sname} &  57.3  & 60.6  & \textbf{62.4} & \textbf{63.1} & \textbf{63.9}   \\
\bottomrule
\end{tabular}

}
\caption{
Performance on GSM8k over training epochs of Llama-3.1-8B when fine-tuned on GSM8k.}
\label{tab:ap_epoch}
\end{table}

\begin{table*}[!htbp]
\centering
\resizebox{1.0\linewidth}{!}{%
\begin{tabular}{c|c|c|c|c | c|c|c|c|c}
\rowcolor[rgb]{.816, .816, .816}
\multicolumn{5}{c|}{\textbf{SAM}} &
\multicolumn{5}{c}{\textbf{L2 regularization}} \\
\toprule
\textbf{$\alpha$} & \textbf{Math (3)} & \textbf{Knowl. (4)} & \textbf{Reason. (2)} & \textbf{Helpful. (2)}
& \textbf{$\lambda$} & \textbf{Math (3)} & \textbf{Knowl. (4)} & \textbf{Reason. (2)} & \textbf{Helpful. (2)} \\
\midrule
0.001 & 52.4  & 73.1  & 65.1  & 43.9
& 0.01   & 53.6& 73.0&65.5 & 43.3  \\
0.01  & 53.3  & 73.3  & 65.5  & 44.0 
& 0.05     &53.1 & 72.2& 64.7 & 42.1 \\
0.1   & 49.4  & 63.9  & 58.4  & 43.4
& 0.1    & 51.6& 70.1 &62.1 & 40.9  \\
\midrule
\rowcolor[rgb]{.816, .816, .816}
\multicolumn{5}{c|}{\textbf{Logit-level consistency}} &
\multicolumn{5}{c}{\textbf{Ensembling}} \\
\toprule
\textbf{$\lambda$} & \textbf{Math (3)} & \textbf{Knowl. (4)} & \textbf{Reason. (2)} & \textbf{Helpful. (2)}
& \textbf{$\beta$} & \textbf{Math (3)} & \textbf{Knowl. (4)} & \textbf{Reason. (2)} & \textbf{Helpful. (2)} \\
\midrule
0.1   & 53.1  & 73.1  & 65.2  & 44.3
& 0.1 & 53.4& 73.3& 64.9 & 44.3 \\
1.0   & 53.9  & 73.2  & 65.2  & 44.9 
&  0.5    &53.1 &	73.4&	65.4	&44.3 \\
10.0    & 52.3  & 69.2  & 62.8  & 42.6
&  0.9   & 40.6	&73.1	&66.0	&44.0\\
\midrule
\rowcolor[rgb]{.816, .816, .816}
\multicolumn{5}{c|}{\textbf{L1 regularization}} &
\multicolumn{5}{c}{\textbf{EMA}} \\
\toprule
\textbf{$\lambda$} & \textbf{Math (3)} & \textbf{Knowl. (4)} & \textbf{Reason. (2)} & \textbf{Helpful. (2)}
& \textbf{$\beta$} & \textbf{Math (3)} & \textbf{Knowl. (4)} & \textbf{Reason. (2)} & \textbf{Helpful. (2)} \\
\midrule

0.01   & 53.1  & 72.2 & 63.6 & 43.4
& 0.1 & 53.3& 71.0& 61.0 & 44.2 \\
0.05     & 50.6 & 59.7 & 65.9 & 43.2
&0.5     & 53.2 & 71.2& 60.9 &44.2 \\
0.1    & 47.8 & 46.5& 50.5 & 42.9
& 0.9    & 53.4  & 71.6  & 61.3 & 44.4 \\
\bottomrule
\end{tabular}
}
\caption{{{Results of hyperparameter searches for ablative baselines, \viz, SAM, Logit-level consistency, L1/L2 regularization, Ensembling, and EMA, using Llama-3.1-8B fine-tuned on GSM8k, evaluated across 11 tasks in four categories.}}}
\label{tab:hyperparameter}
\end{table*}

\begin{table}[t]
\centering
\resizebox{1.0\linewidth}{!}{%
\begin{tabular}{r|c|c|c|c}
\multicolumn{1}{c|}{\textbf{$\lambda$}} & \textbf{Math (3)} & \textbf{Knowl. (4)} & \textbf{Reason. (2)} & \textbf{Helpful. (2)} \\
\midrule
    0.1   & 48.1  & 73.2  & 65.4  & 44.2  \\
    1.0     & 48.1  & 73.4  & 65.4  & 44.2   \\
    5.0     & 54.2  & 73.5  & 65.6  & 45.7   \\
    10.0    & 53.4  & 73.0    & 65.0    & 44.4 \\
    50.0    & 52.5  & 70.1  & 63.1  & 44.6   \\
\bottomrule
\end{tabular}
}
\caption{Results of varying $\lambda$ (with fixed $\alpha = 0.1$) for Llama-3.1-8B fine-tuned on GSM8k, evaluated across 11 tasks in four categories.}
\label{tab:ablation_lbd}
\end{table}

\paragraph{Fine-tuning Qwen3 on GSM8k}
{We further evaluate Qwen3 \cite{qwen3technicalreport} fine-tuned on GSM8k, and the results are shown in Table~\ref{tab:qwen3}, where LfU consistently ranks highest among baselines. While SFT exhibits performance degradation in the reasoning and helpfulness categories, LfU maintains stable performance without such degradation across all categories. This demonstrates that LfU generalizes well across diverse models.}

\paragraph{Computational costs on larger models}
\label{ap:compuation_costs}
{We assess the scalability of our approach by comparing computation costs across different model sizes. As shown in Table~\ref{tab:ap_comp}, both SFT and LfU exhibit a similar growth ratio when scaling from 8B to 13B, indicating that LfU maintains comparable computational scalability to SFT even for larger models.
}

\paragraph{Detailed per-dataset results}
For the detailed task-wise results, Tables~\ref{tab:gsm8k_all}, \ref{tab:arc_all}, \ref{tab:dolly_all}, and \ref{tab:lima_all} present the corresponding performances for GSM8k, ARC-Challenge, Alpagasus Dolly 3k, and LIMA, respectively.

\section{Additional ablation study}
\label{ap:ablation}
\paragraph{Alternative parameter-level averaging designs}

To compare \sname with parameter-level averaging, we consider 
(a) ensembling that interpolates the vanilla model parameters $\vtheta_{\text{vanilla}}$ and the SFT-trained parameters $\vtheta_{\text{SFT}}$, and (b) EMA that updates a smoothed copy $\vtheta_{\text{EMA}}$ toward the current parameters $\vtheta_{\text{current}}$ of the model at each step using coefficient $\beta$. Both methods operate purely on parameters and yield only minor gains over SFT, whereas \sname regularizes representations and achieves clearly superior results (Table~\ref{tab:parameter}).

\paragraph{Impact on vanilla SFT}
\label{ap:impact_on_the_vanilla_SFT}
To examine the impact of our consistency term on the vanilla SFT, we compare their performances of Llama-3.1-8B fine-tuned on GSM8k at each of the training epochs. As shown in Table~\ref{tab:ap_epoch}, while LfU slightly slows down SFT during the early stage (up to epoch 1), it soon achieves higher performance after epoch 2, demonstrating improved data efficiency.

\paragraph{Hyperparameter ablations}
{Additional ablation results for various values of $\lambda$ with fixed $\alpha = 0.1$ are shown in Table~\ref{tab:ablation_lbd}. We observe that overly leveraging the regularization eventually reduces in-domain performance, indicating that it can interfere with SFT. Meanwhile, the results for different $\alpha$ values with fixed $\lambda = 5$ are reported in Table~\ref{tab:ablation_alpha}, showing that a small but non-trivial amount of corruption is necessary for gains, which supports the design principle of \sname.}
Furthermore, the results of SAM \cite{foret2020sharpness}, logit-level consistency, L1/L2 regularization, Ensembling, and EMA with varying hyperparameters are all summarized in Table~\ref{tab:hyperparameter}. These results collectively show that \sname consistently outperforms other approaches across diverse settings.

\begin{table}[t]
\centering
\resizebox{1.0\linewidth}{!}{%
\begin{tabular}{l|c|c|c|c}
\multicolumn{1}{c|}{\textbf{$\alpha$}} & \textbf{Math (3)} & \textbf{Knowl. (4)} & \textbf{Reason. (2)} & \textbf{Helpful. (2)} \\
\midrule
    0.001 & 52.5  & 73.1  & 65.1  & 44.0   \\
    0.01  & 53.0    & 73.1  & 65.2  & 44.3  \\
    0.05 &53.9  & 73.3  & 65.7  & 45.0 \\
    0.1   & 54.2  & 73.5  & 65.6  & 45.7 \\
    0.5   & 44.5  & 60.4  & 54.1  & 43.4   \\
\bottomrule
\end{tabular}
}
\caption{Results of varying $\alpha$ (with fixed $\lambda = 5$)  for Llama-3.1-8B fine-tuned on GSM8k, evaluated across 11 tasks in four categories.}
\label{tab:ablation_alpha}
\end{table}

\begin{table*}[t]
\centering
\resizebox{1.0\linewidth}{!}{%
\begin{tabular}{l|ccc|cccc|cc|cc}
& \multicolumn{3}{c|}{\textbf{Math}} 
& \multicolumn{4}{c|}{\textbf{Knowledge}} 
& \multicolumn{2}{c|}{\textbf{Reasoning}} 
& \multicolumn{2}{c}{\textbf{Helpfulness}} \\
\cmidrule(lr){2-4} \cmidrule(lr){5-8} \cmidrule(lr){9-10} \cmidrule(l){11-12}
\multicolumn{1}{c|}{} 
& \textbf{GSM8k} & \textbf{MathQA} & \textbf{ASDiv} 
& \textbf{MMLU} & \textbf{PIQA} & \textbf{HellaSwag} & \textbf{LAMBADA} 
& \textbf{ARC} & \textbf{CoQA} 
& \textbf{ToxiGen} & \textbf{TruthfulQA} \\
\midrule
\rowcolor[rgb]{.816, .816, .816} \multicolumn{12}{l}{\textbf{Llama-3.1-8B}} \\
\midrule
Vanilla & 26.7  & 39.4  & 47.5  & 63.4  & 81.3  & 78.9  & 68.1  & 51.5  & 80.6  & 42.7  & 45.2 \\
\midrule
SFT & 61.6  & 33.9  & 59.1  & 63.5  & 81.4  & 79.5  & 68.9  & 50.3  & 79.9  & 43.1  & 45.1 \\
NEFTune & 62.1  & 33.1  & 65.1  & 63.0     & 81.6  & 79.5  & 68.9  & 50.3  & 80.2  & 43.3  & 45.3 \\
GEM & 62.6  & 34.1  & 60.6  & 63.9  & 81.0     & 79.7  & 67.5  & 50.7  & 79.8  & 43.2  & 47.0  \\
IM & 57.6  & 37.2  & 60.5  & 62.9  & 81.1  & 79.5  & 68.7  & 51.7  & 79.4  & 43.1  & 45.6 \\
SDFT & 56.6  & 36.2  & 64.4  & 62.5  & 81.6  & 79.8  & 69.5  & 50.3  & 79.7  & 43.2  & 41.0  \\
\midrule
\textbf{\sname} (RepS) &56.3  & 36.0    & 67.3  & 67.4  & 80.8  & 79.1  & 65.7  & 52.0     & 79.1  & 45.2  & 51.8 \\
\textbf{\sname} (LoRA) & 63.1  & 34.2  & 65.2  & 62.6  & 81.9  & 79.5  & 69.8  & 51.2  & 80.0     & 47.1  & 44.2 \\
\midrule
\rowcolor[rgb]{.816, .816, .816} \multicolumn{12}{l}{\textbf{Llama-3.1-8B-Instruct}} \\
\midrule
Vanilla & 62.9  & 39.4  & 79.2  & 69.0    & 81.0    & 79.3  & 66.8  & 83.2  & 78.7  & 84.7  & 55.1 \\
\midrule
SFT & 80.8  & 35.9  & 80.4  & 68.3  & 80.8  & 79.1  & 66.1  & 82.7  & 78.3  & 82.9  & 52.2 \\
NEFTune & 80.3  & 37.1  & 80.7  & 68.3  & 80.6  & 79.2  & 66.4  & 83.0     & 78.4  & 83.1  & 51.5 \\
GEM &  79.2  & 35.3  & 80.5  & 68.3  & 81.0     & 79.9  & 65.8  & 83.1  & 79.5  & 82.3  & 52.9 \\
IM &   76.0    & 36.4  & 79.3  & 67.6  & 81.0     & 79.3  & 68.6  & 83.2  & 79.8  & 84.5  & 52.5 \\
SDFT & 81.2  & 36.0     & 79.4  & 68.4  & 80.7  & 78.9  & 67.8  & 82.1  & 77.4  & 85.0     & 49.8 \\
\midrule
\textbf{\sname} (RepS) &81.3  & 36.0     & 80.1  & 68.5  & 81.0     & 79.3  & 66.2  & 82.8  & 79.1  & 83.3  & 51.8 \\
\textbf{\sname} (LoRA) &82.4  & 36.4  & 81.2  & 68.7  & 80.7  & 79.2  & 67.5  & 83.9  & 79.1  & 83.8  & 54.1 \\
\midrule
\rowcolor[rgb]{.816, .816, .816} \multicolumn{12}{l}{\textbf{Llama-2-7B}} \\
\midrule
Vanilla & 5.2   & 28.1  & 18.1  & 41.8  & 79.1  & 76.0    & 68.5  & 43.4  & 77.4  & 42.9  & 39.0 \\
\midrule
SFT & 29.9  & 27.0     & 50.8  & 39.5  & 78.8  & 76.2  & 69.0     & 42.5  & 77.5  & 42.4  & 38.6 \\
NEFTune &  31.1  & 26.8  & 50.5  & 40.1  & 79.0     & 76.4  & 69.4  & 42.6  & 77.6  & 42.1  & 38.6 \\
GEM &30.0     & 26.7  & 51.1  & 41.4  & 78.7  & 76.4  & 69.0     & 42.4  & 77.4  & 42.2  & 38.9 \\
IM & 20.8  & 27.9  & 48.0     & 39.9  & 79.1  & 76.6  & 69.8  & 44.1  & 77.9  & 42.8  & 36.5 \\
SDFT & 28.0     & 27.0     & 45.7  & 36.8  & 79.3  & 76.4  & 69.8  & 42.8  & 77.2  & 42.6  & 37.5 \\
\midrule
\textbf{\sname} (RepS) & 27.1  & 26.6  & 53.8  & 40.8  & 79.1  & 76.1  & 69.4  & 42.3  & 77.8  & 42.6  & 38.6 \\
\textbf{\sname} (LoRA) & 31.4  & 26.8  & 52.8  & 41.1  & 79.2  & 76.5  & 69.8  & 42.6  & 77.7  & 42.8  & 38.9 \\
\midrule
\rowcolor[rgb]{.816, .816, .816} \multicolumn{12}{l}{\textbf{Mistral-7B-v0.3}} \\
\midrule
Vanilla &7.7   & 16.2  & 20.3  & 59.1  & 82.3  & 80.4  & 69.6  & 48.7  & 81.0    & 42.4  & 42.6 \\
\midrule
SFT & 50.7  & 31.1  & 62.8  & 57.8  & 81.2  & 81.3  & 70.7  & 50.5  & 79.1  & 45.1  & 41.0  \\
NEFTune & 57.5  & 31.4  & 62.1  & 58.0     & 81.4  & 81.4  & 70.8  & 50.7  & 80.1  & 44.3  & 41.7 \\
GEM &54.4  & 32.8  & 62.4  & 57.9  & 82.4  & 81.8  & 70.3  & 51.2  & 79.9  & 43.3  & 42.5 \\
IM &39.3  & 33.0     & 56.7  & 58.7  & 82.0     & 81.3  & 70.4  & 51.2  & 79.5  & 42.4  & 42.2 \\
SDFT & 47.8  & 33.9  & 57.6  & 58.4  & 82.2  & 81.5  & 71.8  & 51.7  & 80.1  & 43.7  & 41.9 \\
\midrule
\textbf{\sname} (RepS) &51.9  & 31.8  & 61.8  & 58.6  & 82.8  & 81.3  & 72.2  & 51.4  & 80.1  & 42.9  & 41.7 \\
\textbf{\sname} (LoRA) & 57.9  & 31.0     & 62.7  & 58.0     & 82.5  & 81.1  & 72.2  & 51.3  & 79.8  & 45.3  & 41.7 \\
\bottomrule
\end{tabular}
}
\caption{Comparison of the performance of language models fine-tuned on GSM8k, evaluated on 11 tasks.}
\label{tab:gsm8k_all}
\end{table*}

\begin{table*}[t]
\centering
\resizebox{1.0\linewidth}{!}{%
\begin{tabular}{l|ccc|cccc|cc|cc}
& \multicolumn{3}{c|}{\textbf{Math}} 
& \multicolumn{4}{c|}{\textbf{Knowledge}} 
& \multicolumn{2}{c|}{\textbf{Reasoning}} 
& \multicolumn{2}{c}{\textbf{Helpfulness}} \\
\cmidrule(lr){2-4} \cmidrule(lr){5-8} \cmidrule(lr){9-10} \cmidrule(l){11-12}
\multicolumn{1}{c|}{} 
& \textbf{GSM8k} & \textbf{MathQA} & \textbf{ASDiv} 
& \textbf{MMLU} & \textbf{PIQA} & \textbf{HellaSwag} & \textbf{LAMBADA} 
& \textbf{ARC} & \textbf{CoQA} 
& \textbf{ToxiGen} & \textbf{TruthfulQA} \\
\midrule
\rowcolor[rgb]{.816, .816, .816} \multicolumn{12}{l}{\textbf{Llama-3.1-8B}} \\
\midrule
Vanilla & 26.7  & 39.4  & 47.5  & 63.4  & 81.3  & 78.9  & 68.1  & 51.5  & 80.6  & 42.7  & 45.2 \\
\midrule
SFT &     10.6  & 40.3  & 59.1  & 63.9  & 81.3  & 79.3  & 68.3  & 80.9  & 80.6  & 44.6  & 45.1 \\
NEFTune &     11.9  & 41.0    & 60.0    & 64.1  & 81.3  & 79.3  & 68.6  & 78.7  & 80.1  & 57.2  & 42.4 \\
GEM &     12.7  & 40.2  & 60.0    & 64.0    & 81.2  & 79.2  & 68.3  & 80.5  & 80.5  & 43.9  & 46.0 \\
IM &     8.3   & 40.6  & 60.0    & 61.7  & 81.1  & 79.4  & 70.5  & 77.2  & 80.9  & 42.8  & 45.3 \\
SDFT &   7.5   & 39.1  & 59.2  & 64.0    & 79.9  & 78.3  & 69.3  & 76.9  & 81.6  & 41.8  & 40.8 \\
\midrule
\textbf{\sname} (RepS) &    10.8  & 41.2  & 60.2  & 61.4  & 81.9  & 80.7  & 70.4  & 79.9  & 80.8  & 42.7  & 45.5 \\
\textbf{\sname} (LoRA) &     11.7  & 40.9  & 61.0    & 64.2  & 80.7  & 79.4  & 68.8  & 80.1  & 80.9  & 45.3  & 45.1 \\
\midrule
\rowcolor[rgb]{.816, .816, .816} \multicolumn{12}{l}{\textbf{Llama-3.1-8B-Instruct}} \\
\midrule
Vanilla & 62.9  & 39.4  & 79.2  & 69.0    & 81.0    & 79.3  & 66.8  & 83.2  & 78.7  & 84.7  & 55.1 \\
\midrule
SFT &     61.7  & 39.7  & 81.0    & 68.9  & 80.9  & 79.0    & 67.2  & 84.3  & 78.4  & 85.1  & 55.1 \\
NEFTune &    62.5  & 40.5  & 80.5  & 69.0    & 80.8  & 78.8  & 67.0    & 84.5  & 78.6  & 84.8  & 56.1 \\
GEM &      61.0    & 38.9  & 81.3  & 68.9  & 80.9  & 79.0    & 67.2  & 84.1  & 80.2  & 85.3  & 55.6 \\
IM &      62.7  & 40.2  & 80.3  & 65.5  & 81.0    & 78.9  & 68.1  & 84.4  & 79.4  & 84.4  & 51.6 \\
SDFT & 61.6  & 38.2  & 80.2  & 68.9  & 79.3  & 76.6  & 68.3  & 80.9  & 78.0    & 84.3  & 50.3 \\
\midrule
\textbf{\sname} (RepS) &60.9  & 41.2  & 81.8  & 68.7  & 81.3  & 78.9  & 67.7  & 84.9  & 78.6  & 85.4  & 55.1 \\
\textbf{\sname} (LoRA) &61.0    & 41.0    & 82.0    & 68.9  & 81.0    & 78.8  & 66.8  & 85.1  & 78.8  & 85.5  & 55.6 \\
\midrule
\rowcolor[rgb]{.816, .816, .816} \multicolumn{12}{l}{\textbf{Llama-2-7B}} \\
\midrule
Vanilla & 5.2   & 28.1  & 18.1  & 41.8  & 79.1  & 76.0    & 68.5  & 43.4  & 77.4  & 42.9  & 39.0 \\
\midrule
SFT & 5.2   & 29.0    & 40.0    & 49.8  & 79.3  & 75.7  & 68.1  & 62.5  & 76.5  & 42.3  & 37.4 \\
NEFTune & 6.0     & 28.9  & 40.0   & 49.6  & 79.2  & 75.8  & 68.1  & 61.5  & 76.9  & 42.6  & 37.0 \\
GEM &5.0     & 29.1  & 38.6  & 50.2  & 79.3  & 75.6  & 68.2  & 64.4  & 76.6  & 42.3  & 38.6 \\
IM &5.0     & 29.2  & 38.9  & 45.1  & 78.5  & 76.4  & 69.8  & 60.9  & 77.7  & 43.5  & 37.5 \\
SDFT & 4.3   & 27.3  & 37.6  & 36.6  & 78.8  & 75.8  & 68.8  & 61.1  & 77.1  & 42.7  & 38.8 \\
\midrule
\textbf{\sname} (RepS) & 5.0     & 28.7  & 42.4  & 50.1  & 79.3  & 75.6  & 68.2  & 64.0    & 76.3  & 43.4  & 37.9 \\
\textbf{\sname} (LoRA) & 5.2   & 28.7  & 43.1  & 50.2  & 79.7  & 75.7  & 68.8  & 63.4  & 76.8  & 43.2  & 37.8 \\
\midrule
\rowcolor[rgb]{.816, .816, .816} \multicolumn{12}{l}{\textbf{Mistral-7B-v0.3}} \\
\midrule
Vanilla &7.7   & 16.2  & 20.3  & 59.1  & 82.3  & 80.4  & 69.6  & 48.7  & 81.0    & 42.4  & 42.6 \\
\midrule
SFT & 8.8   & 35.9  & 16.1  & 61.0    & 81.6  & 80.2  & 69.5  & 79.7  & 78.9  & 43.1  & 42.9 \\
NEFTune & 8.9   & 35.3  & 22.9  & 60.5  & 81.6  & 79.7  & 68.9  & 79.3  & 79.4  & 43.9  & 44.1 \\
GEM &9.2   & 35.7  & 18.2  & 60.9  & 82.6  & 81.1  & 69.4  & 79.4  & 80.0    & 42.9  & 42.6 \\
IM &9.3   & 36.1  & 18.2  & 59.4  & 81.4  & 81.0    & 71.4  & 72.7  & 80.8  & 45.7  & 43.5 \\
SDFT & 8.3   & 35.1  & 19.3  & 61.1  & 81.6  & 79.9  & 69.9  & 78.5  & 79.5  & 43.1  & 42.2 \\
\midrule
\textbf{\sname} (RepS) &8.6   & 35.9  & 20.2  & 61.4  & 81.9  & 80.7  & 70.4  & 79.6  & 79.8  & 45.7  & 43.5 \\
\textbf{\sname} (LoRA) & 9.3   & 34.2  & 20.4  & 61.3  & 81.4  & 80.1  & 69.8  & 79.1  & 79.5  & 46.8  & 43.2 \\
\bottomrule
\end{tabular}
}
\caption{Comparison of the performance of language models fine-tuned on ARC-Challenge, evaluated on 11 tasks.}
\label{tab:arc_all}
\end{table*}

\begin{table*}[t]
\centering
\resizebox{1.0\linewidth}{!}{%
\begin{tabular}{l|ccc|cccc|cc|cc}
& \multicolumn{3}{c|}{\textbf{Math}} 
& \multicolumn{4}{c|}{\textbf{Knowledge}} 
& \multicolumn{2}{c|}{\textbf{Reasoning}} 
& \multicolumn{2}{c}{\textbf{Helpfulness}} \\
\cmidrule(lr){2-4} \cmidrule(lr){5-8} \cmidrule(lr){9-10} \cmidrule(l){11-12}
\multicolumn{1}{c|}{} 
& \textbf{GSM8k} & \textbf{MathQA} & \textbf{ASDiv} 
& \textbf{MMLU} & \textbf{PIQA} & \textbf{HellaSwag} & \textbf{LAMBADA} 
& \textbf{ARC} & \textbf{CoQA} 
& \textbf{ToxiGen} & \textbf{TruthfulQA} \\
\midrule
\rowcolor[rgb]{.816, .816, .816} \multicolumn{12}{l}{\textbf{Llama-3.1-8B}} \\
\midrule
Vanilla & 26.7  & 39.4  & 47.5  & 63.4  & 81.3  & 78.9  & 68.1  & 51.5  & 80.6  & 42.7  & 45.2 \\
\midrule
SFT &29.5  & 40.2  & 41.3  & 63.8  & 80.7  & 78.7  & 67.9  & 50.5  & 79.1  & 43.1  & 58.0 \\
NEFTune & 33.9  & 40.1  & 42.9  & 63.3  & 80.5  & 78.5  & 68.0    & 50.1  & 78.3  & 43.0    & 58.1 \\
GEM & 26.2  & 40.9  & 46.3  & 63.7  & 81.4  & 79.6  & 66.4  & 51.6  & 78.5  & 42.8  & 58.8 \\
IM &  26.2  & 40.6  & 48.3  & 63.6  & 81.4  & 79.5  & 68.7  & 53.5  & 80.3  & 43.8  & 53.4 \\
SDFT & 13.6  & 39.5  & 59.4  & 63.4  & 81.2  & 79.0    & 69.0    & 51.5  & 77.8  & 43.3  & 48.7 \\
\midrule
\textbf{\sname} (RepS) & 27.4  & 40.3  & 50.2  & 64.0    & 81.1  & 78.9  & 67.9  & 50.7  & 79.3  & 43.4  & 58.0 \\
\textbf{\sname} (LoRA)& 34.5  & 40.6  & 54.4  & 63.6  & 81.7  & 78.9  & 68.9  & 51.5  & 78.3  & 43.2  & 58.4 \\
\midrule
\rowcolor[rgb]{.816, .816, .816} \multicolumn{12}{l}{\textbf{Llama-3.1-8B-Instruct}} \\
\midrule
Vanilla & 62.9  & 39.4  & 79.2  & 69.0    & 81.0    & 79.3  & 66.8  & 83.2  & 78.7  & 84.7  & 55.1 \\
\midrule
SFT & 52.6  & 37.5  & 79.7  & 68.2  & 79.2  & 78.2  & 65.9  & 83.2  & 80.3  & 85.4  & 53.2 \\
NEFTune & 55.2  & 37.8  & 80.1  & 68.3  & 79.1  & 78.0    & 65.9  & 83.3  & 80.2  & 85.2  & 53.7 \\
GEM &59.4  & 37.7  & 79.9  & 68.5  & 79.9  & 78.8  & 65.6  & 83.6  & 80.1  & 85.5  & 51.9 \\
IM & 61.1  & 37.1  & 80.7  & 67.1  & 81.1  & 78.7  & 67.4  & 83.7  & 80.1  & 85.0    & 52.5 \\
SDFT & 61.8  & 38.0    & 79.9  & 68.6  & 79.8  & 78.1  & 66.1  & 82.8  & 79.6  & 84.8  & 54.0 \\
\midrule
\textbf{\sname} (RepS) & 55.6  & 37.2  & 79.9  & 68.9  & 79.5  & 78.2  & 66.2  & 83.6  & 80.8  & 85.6  & 52.9 \\
\textbf{\sname} (LoRA) & 61.5  & 38.8  & 80.5  & 68.9  & 79.9  & 78.5  & 67.0    & 83.8  & 80.0    & 85.1  & 53.9 \\
\midrule
\rowcolor[rgb]{.816, .816, .816} \multicolumn{12}{l}{\textbf{Llama-2-7B}} \\
\midrule
Vanilla & 5.2   & 28.1  & 18.1  & 41.8  & 79.1  & 76.0    & 68.5  & 43.4  & 77.4  & 42.9  & 39.0 \\
\midrule
SFT & 5.5   & 29.8  & 36.6  & 44.5  & 78.4  & 77.0    & 67.8  & 43.8  & 70.8  & 42.6  & 49.6 \\
NEFTune & 6.2   & 29.7  & 37.0    & 44.2  & 78.8  & 77.2  & 68.0    & 43.6  & 69.5  & 42.7  & 48.0 \\
GEM &5.6   & 29.2  & 38.1  & 44.4  & 78.6  & 77.7  & 67.4  & 43.2  & 71.3  & 42.7  & 48.8 \\
IM &5.5   & 28.9  & 21.0    & 40.1  & 79.4  & 76.6  & 69.3  & 45.4  & 78.5  & 42.9  & 39.4 \\
SDFT & 5.3 & 29.6  & 3.4   & 41.6  & 78.9  & 76.2  & 67.8  & 43.5  & 76.4  & 42.1  & 43.9 \\
\midrule
\textbf{\sname} (RepS) & 5.7   & 29.3  & 36.4  & 44.6  & 78.9  & 77.3  & 68.0    & 44.0    & 70.5  & 42.7  & 49.3 \\
\textbf{\sname} (LoRA) & 5.7   & 29.8  & 38.0    & 44.5  & 78.6  & 77.2  & 67.7  & 44.6  & 70.9  & 42.3  & 49.8 \\
\midrule
\rowcolor[rgb]{.816, .816, .816} \multicolumn{12}{l}{\textbf{Mistral-7B-v0.3}} \\
\midrule
Vanilla &7.7   & 16.2  & 20.3  & 59.1  & 82.3  & 80.4  & 69.6  & 48.7  & 81.0    & 42.4  & 42.6 \\
\midrule
SFT &10.9  & 36.1  & 39.7  & 59.1  & 82.1  & 80.9  & 69.6  & 50.9  & 80.1  & 44.9  & 45.8 \\
NEFTune &14.0    & 35.1  & 47.5  & 59.4  & 82.4  & 80.8  & 69.3  & 50.3  & 80.4  & 46.5  & 49.3 \\
GEM &9.6   & 36.0    & 9.9   & 62.9  & 87.3  & 84.4  & 75.2  & 51.5  & 80.6  & 43.7  & 45.4 \\
IM &11.5  & 35.8  & 23.0    & 59.1  & 82.6  & 81.2  & 69.6  & 52.6  & 81.5  & 43.2  & 45.9 \\
SDFT & 11.9  & 35.6  & 19.3  & 59.6  & 82.4  & 80.8  & 69.2  & 50.7  & 80.9  & 46.0     & 51.0  \\
\midrule
\textbf{\sname} (RepS) & 10.7  & 35.7  & 35.6  & 59.2  & 82.4  & 81.2  & 69.6  & 51.7  & 81.2  & 43.7  & 45.4 \\
\textbf{\sname} (LoRA) &  15.6  & 36.1  & 53.1  & 58.9  & 82.0    & 81.6  & 70.3  & 51.0     & 80.1  & 45.4  & 51.8 \\
\bottomrule
\end{tabular}
}
\caption{Comparison of the performance of language models fine-tuned on  Alpagasus Dolly 3k, evaluated on 11 tasks.}
\label{tab:dolly_all}
\end{table*}

\begin{table*}[t]
\centering
\resizebox{1.0\linewidth}{!}{%
\begin{tabular}{l|ccc|cccc|cc|cc}
& \multicolumn{3}{c|}{\textbf{Math}} 
& \multicolumn{4}{c|}{\textbf{Knowledge}} 
& \multicolumn{2}{c|}{\textbf{Reasoning}} 
& \multicolumn{2}{c}{\textbf{Helpfulness}} \\
\cmidrule(lr){2-4} \cmidrule(lr){5-8} \cmidrule(lr){9-10} \cmidrule(l){11-12}
\multicolumn{1}{c|}{} 
& \textbf{GSM8k} & \textbf{MathQA} & \textbf{ASDiv} 
& \textbf{MMLU} & \textbf{PIQA} & \textbf{HellaSwag} & \textbf{LAMBADA} 
& \textbf{ARC} & \textbf{CoQA} 
& \textbf{ToxiGen} & \textbf{TruthfulQA} \\
\midrule
\rowcolor[rgb]{.816, .816, .816} \multicolumn{12}{l}{\textbf{Llama-3.1-8B}} \\
\midrule
Vanilla & 26.7  & 39.4  & 47.5  & 63.4  & 81.3  & 78.9  & 68.1  & 51.5  & 80.6  & 42.7  & 45.2 \\
\midrule
SFT & 26.9  & 39.6  & 35.0    & 64.2  & 81.2  & 80.2  & 68.4  & 50.9  & 80.0    & 44.0    & 47.4 \\
NEFTune & 26.4  & 39.5  & 35.4  & 64.0    & 81.7  & 80.2  & 68.3  & 50.9  & 80.4  & 43.7  & 47.9 \\
GEM &25.2  & 38.5  & 34.5  & 64.4  & 81.2  & 80.2  & 68.1  & 49.7  & 80.4  & 42.9  & 50.1 \\
IM & 29.1  & 40.1  & 40.1  & 64.2  & 81.8  & 80.4  & 68.0    & 51.9  & 80.6  & 44.1  & 46.9 \\
SDFT & 22.7  & 40.0    & 39.0   & 63.6  & 81.3  & 79.2  & 69.0    & 49.2  & 80.6  & 43.0    & 42.1 \\
\midrule
\textbf{\sname} (RepS) & 27.0    & 39.9  & 39.1  & 64.2  & 81.8  & 80.2  & 68.4  & 50.7  & 80.1  & 43.9  & 47.6 \\
\textbf{\sname} (LoRA) & 25.9  & 40.0    & 38.4  & 64.4  & 81.7  & 80.4  & 68.6  & 51.5  & 80.2  & 43.6  & 47.9 \\
\midrule
\rowcolor[rgb]{.816, .816, .816} \multicolumn{12}{l}{\textbf{Llama-3.1-8B-Instruct}} \\
\midrule
Vanilla & 62.9  & 39.4  & 79.2  & 69.0    & 81.0    & 79.3  & 66.8  & 83.2  & 78.7  & 84.7  & 55.1 \\
\midrule
SFT & 44.8  & 39.0    & 73.5  & 67.1  & 80.4  & 79.3  & 67.9  & 83.1  & 76.8  & 83.7  & 44.9 \\
NEFTune & 48.6  & 38.9  & 73.8  & 67.2  & 80.3  & 79.3  & 67.8  & 82.6  & 77.0    & 84.0    & 45.6 \\
GEM &46.8  & 38.2  & 73.6  & 67.0    & 80.5  & 79.4  & 68.5  & 82.6  & 78.4  & 81.3  & 48.6 \\
IM & 41.1  & 38.8  & 75.6  & 66.5  & 80.6  & 79.1  & 69.0    & 83.1  & 80.7  & 85.7  & 46.6 \\
SDFT &53.8  & 38.4  & 78.4  & 68.3  & 79.7  & 78.7  & 66.5  & 82.4  & 78.7  & 72.3  & 45.3 \\
\midrule
\textbf{\sname} (RepS) &49.3  & 39.2  & 74.2  & 67.2  & 80.3  & 79.4  & 68.0    & 82.7  & 78.6  & 84.1  & 46.1 \\
\textbf{\sname} (LoRA) &52.9  & 37.5  & 74.9  & 67.9  & 80.7  & 78.7  & 67.5 & 83.5  & 77.9  & 85.3  & 46.8 \\
\midrule
\rowcolor[rgb]{.816, .816, .816} \multicolumn{12}{l}{\textbf{Llama-2-7B}} \\
\midrule
Vanilla & 5.2   & 28.1  & 18.1  & 41.8  & 79.1  & 76.0    & 68.5  & 43.4  & 77.4  & 42.9  & 39.0 \\
\midrule
SFT & 5.5   & 29.1  & 17.5  & 40.6  & 79.2  & 76.8  & 69.3  & 43.4  & 76.8  & 41.6  & 40.2 \\
NEFTune & 5.8   & 29.1  & 17.1  & 40.8  & 79.1  & 77.0    & 69.4  & 43.1  & 76.5  & 41.7  & 40.2 \\
GEM & 6.4   & 28.1  & 16.7  & 41.3  & 77.8  & 77.0    & 69.7  & 43.5  & 76.6  & 42.1  & 44.7 \\
IM & 5.5   & 28.6  & 15.7  & 40.8  & 78.6  & 76.4  & 68.2  & 44.7  & 78.3  & 42.9  & 38.1 \\
SDFT & 5.4   & 28.9  & 16.6  & 40.0    & 79.2  & 77.5  & 69.5  & 42.2  & 76.3  & 42.1  & 40.8 \\
\midrule
\textbf{\sname} (RepS) & 6.1   & 29.2  & 22.9  & 41.0    & 79.1  & 76.9  & 69.5  & 43.5  & 76.6  & 41.3  & 40.4 \\
\textbf{\sname} (LoRA) & 6.5   & 29.0    & 16.5  & 41.6  & 79.0    & 77.0    & 68.9  & 43.1  & 77.6  & 42.1  & 41.3 \\
\midrule
\rowcolor[rgb]{.816, .816, .816} \multicolumn{12}{l}{\textbf{Mistral-7B-v0.3}} \\
\midrule
Vanilla &7.7   & 16.2  & 20.3  & 59.1  & 82.3  & 80.4  & 69.6  & 48.7  & 81.0    & 42.4  & 42.6 \\
\midrule
SFT & 13.9  & 34.6  & 49.6  & 59.6  & 82.1  & 82.2  & 70.7  & 50.4  & 80.5  & 45.7  & 46.7 \\
NEFTune & 14.6  & 34.4  & 51.2  & 60.3  & 82.5  & 82.4  & 70.5  & 50.6  & 80.3  & 46.8  & 47.9 \\
GEM & 11.7  & 34.5  & 50.1  & 60.2  & 81.4  & 82.7  & 70.6  & 50.1  & 80.2  & 46.6  & 48.4 \\
IM & 14.2  & 35.2  & 25.2  & 58.6  & 81.9  & 81.1  & 69.3  & 51.6  & 80.5  & 44.1  & 46.3 \\
SDFT & 14.1  & 34.9  & 45.7  & 59.0    & 82.4  & 82.6  & 69.7  & 49.7  & 79.6  & 42.9  & 49.6 \\
\midrule
\textbf{\sname} (RepS) & 14.3  & 35.0    & 51.8  & 59.2  & 82.0    & 82.6  & 69.9  & 49.7  & 81.7  & 44.7  & 46.1 \\
\textbf{\sname} (LoRA) & 14.3  & 35.3  & 50.5  & 59.6  & 82.3  & 82.7  & 70.9  & 51.2  & 80.9  & 45.3  & 47.6 \\
\bottomrule
\end{tabular}

}
\caption{Comparison of the performance of language models fine-tuned on LIMA, evaluated on 11 tasks.}
\label{tab:lima_all}
\end{table*}

\end{document}